%% file: neurips_2026.tex
\documentclass{article}
\usepackage{amsmath}

 \usepackage[main, final]{neurips_2026}

\usepackage[utf8]{inputenc} 
\usepackage[T1]{fontenc}    
\usepackage{hyperref}       
\usepackage{url}            
\usepackage{booktabs}       
\usepackage{amsfonts}       
\usepackage{nicefrac}       
\usepackage{microtype}      
\usepackage{xcolor}         
\usepackage{subcaption}
\usepackage{titletoc}
\newcommand\DoToC{%
  \startcontents
  \printcontents{}{1}{\textbf{Table of Contents}\vskip3pt\hrule\vskip5pt}
  \vskip3pt\hrule\vskip5pt
}

\title{Conformal Prediction for \\Distribution-to-Distribution Regression}

\usepackage[utf8]{inputenc} 
\usepackage[T1]{fontenc}    
\usepackage{hyperref}       
\usepackage{url}            
\usepackage{tabularx}        
\usepackage{algorithm} 
\usepackage{algorithmicx} 
\usepackage{algcompatible} 

\usepackage{algpseudocode} 

\usepackage{amsfonts}       
\usepackage{nicefrac}       
\usepackage{microtype}      
\usepackage{xcolor}         

\usepackage{subcaption}

\usepackage{wrapfig}
\usepackage{amsfonts}
\usepackage{amssymb}
\usepackage{bm}
\usepackage{amsmath}
\usepackage{amsthm}
\usepackage{stmaryrd}
\usepackage{color}
\usepackage{xcolor}
\usepackage[subnum]{cases}
\usepackage{rotating}
\usepackage{enumitem}
\usepackage{accents}
\usepackage[title]{appendix}
\usepackage{mathtools}
\usepackage{caption}
\usepackage{url}
\usepackage{rotating}
\usepackage{framed}
\usepackage{lscape}
\usepackage{multirow}
\usepackage{latexsym}
\usepackage{mathrsfs}
\usepackage[new]{old-arrows}
\usepackage{array}
\usepackage{makecell}
\usepackage{float}
\usepackage{tabularray}
\usepackage{tablefootnote}
\usepackage{tikz}
\usepackage[all]{xy}
\usepackage{tikz-cd}
\usepackage[usestackEOL]{stackengine}
\usepackage{pdflscape}
\usepackage{autobreak}
\usepackage{comment}
\usepackage{booktabs}
\usepackage{graphicx}

\theoremstyle{plain}
\newtheorem{theorem}{Theorem}[section]

\newtheorem{corollary}[theorem]{Corollary}
\newtheorem{assumption}[theorem]{Assumption}

\theoremstyle{definition}

\newtheorem{remark}[theorem]{Remark}

\usepackage{adjustbox}
\usepackage{booktabs}
\usepackage{siunitx}
\usepackage{listings}
\usepackage{colortbl}

\author{%
  Trung-Khang Tran\thanks{These authors contributed equally. Correspondence to: \texttt{ttk0106@u.nus.edu} \& \texttt{tanmn@nus.edu.sg}.}  \\
  School of Computing\\
  National University of Singapore\\
  \texttt{ttk0106@u.nus.edu} \\
  \And
  Tuan A. Hoang\footnotemark[1]  \\
  Center for AI Research\\
  VinUniversity\\
  \texttt{tuan.ha3@vinuni.edu.vn} \\
  \And
  Viet-Hoang Tran\footnotemark[1]  \\
  Department of Mathematics\\
  National University of Singapore\\
  \texttt{hoang.tranviet@u.nus.edu} \\
  \And
  Tan M. Nguyen \\
  Department of Mathematics\\
  National University of Singapore\\
  \texttt{tanmn@nus.edu.sg} \\
}

\usepackage[capitalize,noabbrev]{cleveref}
\definecolor{refcolor}{HTML}{000099}
\hypersetup{ %
pdftitle={},
pdfauthor={},
pdfsubject={},
pdfkeywords={},
pdfborder=0 0 0,
pdfpagemode=UseNone,
colorlinks=true,
linkcolor=refcolor,
citecolor=refcolor,
filecolor=refcolor,
urlcolor=refcolor,    
pdfview=FitH}

\begin{document}

\maketitle

\begin{abstract}
  Conformal Prediction (CP) provides a model-agnostic framework for constructing finite-sample prediction sets with coverage guarantees and has become a tool for reliable decision-making in high-risk settings. Recent advances have extended Split Conformal Prediction to increasingly complex learning problems, including multilabel, high-dimensional, and functional outputs. In this work, we study CP for Distribution-to-Distribution Regression, a setting of growing interest across many application domains. We show that a naive application of Split CP yields prediction sets that, while marginally valid, are non-interpretable, e.g., lacking a well-defined volume or a tractable sampling procedure, and of limited practical value. To overcome this, we propose a new conformal framework that operates in an orthogonal basis space, producing interpretable prediction sets while preserving coverage guarantees. We further develop an adaptive variant with asymptotic conditional coverage under mild assumptions. Empirical results on synthetic and real-world datasets validate the effectiveness of the proposed approach. The code is publicly available at {\hypersetup{urlcolor=blue}\urlstyle{tt}%
\url{https://github.com/tuanhacs/conformal_d2d}}.
\end{abstract}


\input{sections/1_introduction}

\input{sections/2_Preliminaries}

\input{sections/3_Naive_Conformal_for_D2D_and_the_lack_of_interpretation}

\input{sections/4_Conformal_Prediction_for_D2D}

\input{sections/5_Adaptive_Conformal_for_Conditional_Guarentee}

\input{sections/6_Experiment}

\bibliography{neurips_2026}
\bibliographystyle{plain}


\clearpage

\appendix

\input{sections/appendix1}


\newpage
\input{checklist.tex}

\end{document}

%% file: sections/1_introduction.tex
\section{Introduction} \label{sec: Intro}







\textit{Conformal Prediction (CP)} \cite{papadopoulos2002inductive,vovk2005algorithmic,lei2014distribution,thurin2025optimal,narteni2025novel,patel2025conformal} is a model-agnostic and distribution-free framework that quantifies predictive uncertainty by constructing prediction sets with rigorous coverage guarantees, ensuring the true outcome lies within the set at a user-specified confidence level. CP has been widely adopted across many domains, including medical imaging \cite{lu2022improving,lu2022fair,  kutiel2023conformal}, classical machine learning \cite{balasubramanian2014conformal, johansson2014regression}, deep learning \cite{papadopoulos2007conformal, wieslander2020deep, zhang2021deep, zargarbashi2023conformal}, diffusion models \cite{teneggi2023trust,sun2023conformal, wang2024dm, huuncertain}, large language models \cite{cherian2024large, vishwakarmaprune}, image segmentation \cite{gade2024impact,mossina2025conformal, chen2025conformalsam, cheung2025compass, yapicioglu2025conformasegment}, and federated learning \cite{lu2021distribution,lemercier2021distribution, lu2023federated, plassier2023conformal, humbert2023one}.

Recent work has extended CP to more complex settings, where uncertainty is inherently greater. Notable examples include conformalized quantile regression \cite{romano2019conformalized}, risk control \cite{angelopoulos2022conformal}, multi-label classification \cite{papadopoulos2014cross,paisios2019deep,tyagi2023multi}, multivariate regression \cite{klein2025multivariate,thurin2025optimal}, and functional data analysis \cite{lei2015conformal}. A key challenge in these extensions is preserving the interpretability of CP sets, particularly their ability to quantify uncertainty through volume and support uniform sampling for downstream tasks. While interpretability can often be maintained within the standard \textit{Split Conformal Prediction} framework \cite{papadopoulos2008inductive}, naive applications may yield poorly interpretable prediction sets in certain settings. One such case is \textit{Distribution-to-Distribution Regression (DDR)} \cite{oliva2013distribution}, where both inputs and outputs are distributions observed only through finite samples. At test time, the goal is to infer an output distribution from samples drawn from an input distribution. Despite extensive study in prior work, including \cite{guestrin2004distributed, klein2014scalable, szabo2016learning, chen2019lqd, song2019distribution, li2021deep, ghodrati2022distribution, okano2023distribution, chen2023wasserstein, chen2024distribution}, the intrinsic complexity of distribution-valued data and limited samples per distribution fundamentally constrain predictive accuracy, limiting reliable uncertainty quantification and interpretability of standard Split Conformal Prediction approaches.

\textbf{Contribution. }Motivated by this gap, we propose a CP framework for the DDR setting that ensures coverage guarantees, thereby enhancing confidence in distributional predictions while preserving the interpretability inherent to CP methods. We further introduce an adaptive variant of the framework that provides asymptotic conditional coverage guarantees
in the DDR setting, enabling valid uncertainty quantification when conditioning on a given input distribution. The paper is organized as follows:
\begin{enumerate}
    \item In \cref{sec: Naive Approac}, we examine a naive Split Conformal Prediction approach for DDR that follows existing frameworks in the literature, and we rigorously demonstrate why such an approach fails to preserve interpretability.
    \item In \cref{sec: Mainmeth}, we present our CP framework for DDR. By projecting distributions into a vector space, we show that the resulting framework ensures coverage guarantees while remaining interpretable.
    \item In \cref{sec: AdapCP}, we propose an adaptive extension of our framework that provides asymptotic coverage guarantees conditional on the input distribution under mild assumptions.
    \item In \cref{sec: Exp}, we report empirical results on both synthetic and real datasets across multiple DDR baselines for both the standard and adaptive CP frameworks. We additionally present experiments illustrating the relationship between coverage and calibration set size, as well as a comparison of inference time between the two CP frameworks. We also study the effect of the low-data regime on our framework.
    
\end{enumerate} 
\cref{sec: Prelim} provides background on CP and the DDR problem. A table of notation, background on CP, theoretical foundations, and additional experimental details are provided in the Appendix.

%% file: sections/2_Preliminaries.tex
\section{Preliminaries} \label{sec: Prelim}

This section provides a brief overview of Conformal Prediction (CP) and establishes the problem formulation and notation for Distribution-to-Distribution Regression (DDR). \textit{We refer to Appendix~\ref{app: backgroundConf} for more details on CP, and Appendix~\ref{app: apliinDDR} for applications of CP in DDR.}

\textbf{Conformal Prediction (CP). }
Let $\hat{f}:\mathcal{X}\to\mathcal{Y}$ be a \textit{prediction model}.
Given a \textit{target miscoverage level} $\gamma\in(0,1)$, CP aims to construct a \textit{set-valued predictor}
$\mathcal{Q}:\mathcal{X}\to\mathcal{P}(\mathcal{Y})$
such that, for a test pair $(X_0,Y_0)\in\mathcal{X}\times\mathcal{Y}$, we obtain the following \textit{coverage guarantee}:
\begin{align}\label{eq: CovGuaren}
\mathbb{P}\!\left(Y_0\in\mathcal{Q}(X_0)\right)\ge 1-\gamma.
\end{align}
A widely used conformal prediction strategy is \textit{Split Conformal Prediction} \cite{papadopoulos2008inductive}. In this framework, the available data are partitioned into a training set and an independent \textit{calibration set}. The predictive model $\hat{f}$ is trained exclusively on the training set, while the calibration set, unseen during training, is used to evaluate a \textit{nonconformity score} that quantifies the discrepancy between the model prediction $ \hat f(X) $ and the observed response $ Y $. In particular, let $d_0$ denote the size of the calibration set, and let 
$\{s_i\}_{i=1}^{d_0}$ be the corresponding nonconformity scores, with 
\begin{align*}
s_i \coloneq s\big(\hat{f}(X^c_i), Y^c_i\big),\quad (X^c_i,Y^c_i) \in \text{Calibration Set}.
\end{align*}
The choice of the score function $s$ is problem-dependent and reflects the data structure and uncertainty relevant to the task \cite{papadopoulos2007conformal,angelopoulos2021gentle, andeol2023confident}. A threshold is then constructed by selecting the empirical quantile
\begin{align*}
\hat q
\coloneqq
\text{the } \dfrac{\left\lceil (d_0+1)(1-\gamma) \right\rceil}{d_0}\text{ quantile of } \{s_i\}_{i=1}^{d_0},
\end{align*}
equivalently, the $ \lceil (d_0+1)(1-\gamma)\rceil $-th order statistic of $\{s_i\}_{i=1}^{d_0}$. The \textit{prediction set} is defined as
\begin{align*}
\mathcal{Q}(X_0)
\coloneqq
\big\{ Y \in \mathcal{Y} : s(\hat f(X_0), Y) \le \hat q \big\},
\end{align*}
which guarantees the marginal coverage in \cref{eq: CovGuaren}. 

\textbf{Distribution-to-Distribution Regression (DDR). }
Let $\mathcal{X}$ denote a class of input distributions supported on  $\Psi^{k} \subseteq \mathbb{R}^{k}$, and let $\mathcal{Y}$ denote a class of output distributions supported on $\Lambda^{l} \subseteq \mathbb{R}^{l}$.
We assume that each distribution in $\mathcal{X}$ and $\mathcal{Y}$ admits a density with respect to the Lebesgue measure.
Importantly, the elements of $\mathcal{X}$ and $\mathcal{Y}$ are not restricted to probability distributions and therefore need not be everywhere nonnegative nor integrate to one. The goal of DDR  is to learn a functional mapping
$
f : \mathcal{X} \to \mathcal{Y}.
$ To estimate $f$, we consider a collection of paired distributions $ \{(X_1, Y_1), \ldots, (X_d, Y_d)\}$, where $X_i \in \mathcal{X}$, $Y_i \in \mathcal{Y}$. Each input distribution is assumed to be sampled independently and identically from a probability measure $\mathcal{P}$ defined on $\mathcal{X}$, that is,
$
X_1, \ldots, X_d \overset{\text{i.i.d.}}{\sim} \mathcal{P}
$. Due to the nature of DDR in real-world settings, the distributions $X_i$ and $Y_i$ cannot be observed directly for all $i \in \{1,\ldots,d\}$. Instead, we observe independent and identically distributed samples from each distribution. Specifically, the available dataset consists of paired input/output samples
$
\mathcal{D} = \{(\bar{M}_1, \bar{N}_1), \ldots, (\bar{M}_d, \bar{N}_d)\},
$
where $\bar{M}_i = \{M_{i1}, \ldots, M_{i m_i}\}$ with $M_{ij} \overset{\text{iid}}{\sim} X_i$, and
$
\bar{N}_i = \{N_{i1}, \ldots, N_{i n_i}\}
\quad \text{with} \quad
N_{ij} \overset{\text{iid}}{\sim} Y_i.
$ Finally, let $\{x_i\}_{i=1}^{d}$ and $\{y_i\}_{i=1}^{d}$ denote the probability density functions corresponding to the distributions $\{X_i\}_{i=1}^{d}$ and $\{Y_i\}_{i=1}^{d}$, respectively.
Throughout this work, we adopt a general, nonparametric setting with minimal assumptions on the data. In particular, we make only weak assumptions on the structure of the input and output distributions and the underlying measure from which the inputs are sampled.

%% file: sections/3_Naive_Conformal_for_D2D_and_the_lack_of_interpretation.tex
\section{Limitations of Naive Split CP for DDR}\label{sec: Naive Approac}
In this section, we analyze a naive extension of Split CP to DDR and show that it fails to preserve the interpretability of the resulting prediction sets. We also discuss why interpretability is essential in CP.

\textbf{Split CP and Interpretability in Classical Settings.}
In most classical applications, CP can be applied directly via the Split CP framework across diverse learning settings and output spaces. Early work studied classification with discrete outputs, where the prediction set $\mathcal Q$ lies in a finite label space \cite{shafer2008tutorial,papadopoulos2011regression,angelopoulos2021gentle}. Later extensions considered one-dimensional regression with real-valued outputs, yielding interval-valued sets \cite{johansson2014regression,romano2019conformalized}. Recent work further extends CP to vector-valued inputs and outputs, producing geometrically structured sets in higher dimensions \cite{klein2025multivariate,thurin2025optimal}. Despite differences in data modality and output structure across these settings, the resulting CP sets remain \textit{interpretable}: Their volume can be explicitly quantified and uniform sampling from $\mathcal Q$ is feasible. This interpretability enables meaningful uncertainty quantification and is essential to CP, as discussed later in this section.

\textbf{Failure of Naive Split CP in DDR. }
When inputs and outputs are further generalized to distribution-valued data, a naive application of Split CP fails to preserve the interpretability properties described above. This occurs because the resulting CP sets reside in infinite-dimensional function spaces, where notions such as volume, geometric structure, and uniform sampling are no longer well defined or computationally tractable.

We now illustrate this issue formally.
Consider a model $f:\mathcal{X}\to\mathcal{Y}$. Following the standard CP procedure, the dataset $\mathcal{D}$ is split into a training set $\mathcal{D}_t$ of size $d_t$ and a calibration set $\mathcal{C}$ of size $d_0$. The model is first trained on $\mathcal{D}_t$ to obtain a fitted predictor $\hat f$. Then, for each calibration pair $(\bar M_i,\bar N_i)\in\mathcal C$, the underlying input and response distributions are approximated through some approximation function $\psi$, yielding
$
\tilde X_i := \psi(\bar M_i),\text{ and }
\tilde Y_i := \psi(\bar N_i).
$
The calibration nonconformity score is then computed as
$
s_i
:=
s\!\left(\hat f(\tilde X_i),\tilde Y_i\right),
$
where $s:\mathcal{Y}\times\mathcal{Y}\to\mathbb{R}$ measures the discrepancy between the predicted and observed outputs. We denote $\hat q$ as 
$
\text{the } \left\lceil (d_0+1)(1-\gamma) \right\rceil/d_0\text{ quantile of } \{s_i\}_{i=1}^{d_0}.
$ Given a test input $\bar M_{0}$ with associated (unobserved) output $\bar N_{0}$, the resulting CP set is
\begin{align}\label{eq: NaiveCon}
\mathcal{Q}_{\text{Naive}}(\bar M_{0})
:=
\big\{
Y\in\mathcal{Y} :
s\big(\hat f(\tilde X_{0}), Y\big)\le \hat q
\big\}.
\end{align}
The naive prediction set in \cref{eq: NaiveCon} still satisfies the marginal coverage guarantee of CP:
\begin{align*}
\mathbb{P}\left(
Y_{0} \in \mathcal{Q}_{\text{Naive}}(\bar M_{0})
\right)\ge 1-\gamma .
\end{align*}
However, the naive CP set $\mathcal{Q}_{\text{Naive}}$ is no longer interpretable in the DDR setting. First, although the supports of $\mathcal{X}$ and $\mathcal{Y}$ lie in Euclidean spaces, the induced spaces of admissible distributions are irregular subsets of infinite-dimensional function spaces. These sets need not be connected or smoothly parametrized, which makes fundamental geometric quantities, such as volume, ill-defined or computationally intractable.
Second, the prediction set $\mathcal{Q}_{\text{Naive}}$ consists of distributions rather than finite- or vector-valued objects. Consequently, sampling an element from $\mathcal{Q}_{\text{Naive}}$ requires sampling an entire distribution, for which no canonical notion of uniformity exists in functional spaces.

\textbf{Interpretability in CP. }
Interpretability is a central component of CP, as prediction sets are intended not only to provide coverage guarantees but also to offer meaningful characterizations of predictive uncertainty. In classical settings, the geometry of the conformal set, such as its size, shape, or volume, serves as a direct measure of uncertainty \cite{liu2025uncertainty}, while the ability to sample uniformly from the set supports practical downstream tasks, including visualization \cite{kutiel2023conformal}, simulation \cite{leroy2021conformal}, and decision-making \cite{vishwakarma2024prune} under uncertainty. When these interpretability properties are lost, conformal prediction sets become difficult to use in practice, even if formal coverage guarantees are retained.


%% file: sections/4_Conformal_Prediction_for_D2D.tex
\section{An Interpretable CP Framework for DDR} \label{sec: Mainmeth}
This section provides a CP framework for the DDR problem that satisfies the coverage guarantee in \cref{eq: CovGuaren}. The framework also yields prediction sets with volume characterization and supports uniform sampling. We further analyze the role of the calibration set size and establish finite-sample results on the variability of conditional coverage. We summarize our method in \cref{alg:cp_ddr} in the Appendix.
\subsection{Construction of Our Framework}
Our construction has four steps: we expand the output distributions in a basis, impose regularity on the resulting coefficients, estimate these coefficients from finite samples, and construct CP sets in coefficient space, then map them back to distributions.

\textbf{Basis Expansion of Output Distributions.}
We begin by specifying a functional representation for distributions on
$\Lambda^l$. To this end, we first fix an orthonormal basis
$\{\beta_i\}_{i \in \mathbb{Z}}$ of $L_2(\Lambda)$. For any integer $l \ge 1$, this
basis naturally induces a tensor-product basis on the product space
$\Lambda^l$. Specifically, we construct the collection
$\{\beta_\alpha\}_{\alpha \in \mathbb{Z}^l}$ by defining
\begin{align*}
\beta_\alpha(z) \coloneqq \textstyle\prod_{i=1}^l \beta_{\alpha_i}(z_i),
\text{ where }~ z = (z_1,\dots,z_l) \in \Lambda^l .
\end{align*}
It is well known that the collection $\{\beta_\alpha\}_{\alpha \in \mathbb{Z}^l}$
forms an orthonormal basis of $L_2(\Lambda^l)$, with inner products satisfying
$
\langle \beta_\alpha, \beta_\gamma \rangle = \mathbb{I}_{\{\alpha = \gamma\}} .
$ We next describe how an arbitrary
distribution in $\mathcal{Y}$ can be expressed in this basis. Given any
$Y \in \mathcal{Y}$ that admits a density $y \in L_2(\Lambda^l)$ with respect to
the Lebesgue measure, the density $y$ admits the orthogonal expansion
$
y(z) \coloneqq \sum_{\alpha \in \mathbb{Z}^l} \theta_\alpha(Y)\, \beta_\alpha(z),  
$ for all $z \in \Lambda^l$, where the expansion coefficients $\theta_\alpha(Y) \in \mathbb{R}$ are given by
\begin{align*}
\theta_\alpha(Y)
\coloneqq  \int_{\Lambda^l} \beta_\alpha(z)\, y(z)\, dz
= \int_{\Lambda^l} \beta_\alpha(z)\, dY(z).
\end{align*}
\textbf{Regularity Assumption.}
We next impose an anisotropic Sobolev ellipsoid assumption on the projection coefficients as stated in Assumption~\ref{ass: A1}.

\begin{assumption} \label{ass: A1}
Assume that the projection coefficients satisfy an anisotropic Sobolev ellipsoid
condition. In particular, let
$
\theta(Y) = \{\theta_\alpha(Y)\}_{\alpha \in \mathbb{Z}^l}
$
denote the collection of projection coefficients associated with a distribution
$Y$. We assume 
$\mathcal{Y} = \{ Y : \theta(Y) \in \Theta(\nu, \sigma, \bar{A}) \},
$ where
\begin{align*}
\Theta(\nu, \sigma, \bar{A})
= \Big\{ \big\{\theta_\alpha \big\}_{\alpha \in \mathbb{Z}^l} :~
\textstyle\sum_{\alpha \in \mathbb{Z}^l} \theta_\alpha^2  w_\alpha^2(\nu, \sigma) < \bar{A} \Big\},
\end{align*}
and the anisotropic weight function is given by
$
w_\alpha^2(\nu, \sigma)
= \textstyle\sum_{i=1}^l (\nu_i |\alpha_i|)^{2\sigma_i}, \text{ where } ~
 \nu_i, \sigma_i, \bar{A} > 0.
$
\end{assumption}
\textbf{Empirical Approximation of Output Distributions.}
In practice, the true distribution $Y_i$ is not directly observed; instead, one
has access only to a finite sample drawn from $Y_i$. Let $\widehat{Y}_i$ denote
the empirical distribution constructed from $n_i$ observations by assigning
equal mass to each data point. Concretely, $\widehat{Y}_i$ is defined such that
$
\widehat{Y}_i(N = N_{ij}) = {1}/{n_i}, \text{ for all} ~~ j = 1, \ldots, n_i.
$
We define an estimator of the density
$y_i$ using a truncated orthogonal series expansion. The truncation serves to
balance the approximation error in accordance with
Assumption~\ref{ass: A1}. Specifically, we define
$
\tilde{y}_i(z) \coloneqq \sum_{\alpha \in T_R}
\theta_\alpha(\widehat{Y}_i)\, \beta_\alpha(z),
$
where the empirical projection coefficients are given by
\begin{align*}
\theta_\alpha(\widehat{Y}_i)
= \int_{\Lambda^l} \beta_\alpha(z)\, d\widehat{Y}_i(z)
= \frac{1}{n_i} \sum_{j=1}^{n_i} \beta_\alpha(N_{ij}).
\end{align*}
Here, the index set
$
T_R = \{ \alpha \in \mathbb{Z}^l : 0 < w_\alpha(\nu, \sigma) \le R \}
$
corresponds to a computational truncation that restricts the expansion to
finitely many basis functions. Under the anisotropic Sobolev ellipsoid assumption and with an appropriate choice of the truncation parameter $R$, the resulting estimator achieves an optimal mean-squared error rate. In particular, it can be
shown that
$
\mathbb{E} \left[ \lVert \tilde{y}_i - y_i \rVert_2^2 \right]
= \mathcal{O}\Big( n_i^{-2/\left(2 + \sum_{j} 1/\sigma_j\right)} \Big),
$ where the rate holds for $n_i \to \infty$ \cite{nussbaum1983optimal}.

\textbf{Construction of Prediction Set.}
We next describe the conformal calibration step. To this end, we partition the
available data and designate a subset as the calibration set,
$
\mathcal{C}
=
\{(X_1^c, Y_1^c), \ldots, (X_{d_0}^c, Y_{d_0}^c)\}.
$
As in the training phase, the latent distributions
$\{(X_i^c, Y_i^c)\}_{i=1}^{d_0}$ are not directly observed. Instead, we observe
finite samples drawn from these distributions, denoted by
$
\{(\bar{M}_1^c, \bar{N}_1^c), \ldots, (\bar{M}_{d_0}^c, \bar{N}_{d_0}^c)\}.
$
Let $\hat{f}$ denote a fitted model obtained from the training data. For each
$i \in [d_0]$, we define our score as
\begin{align*}
s_i:=s(\hat f(X_i^c), Y_i^c) = g\!\left( W^{1/2} \big( \theta(\hat{f}(X_i^c)) - \theta(\tilde{y}_i^c) \big) \right),
\end{align*}
where $g : \mathbb{R}^{|T_R|} \to \mathbb{R}_{\ge 0}$ is a metric on the coefficient space, and $W$ is a weighting matrix that encodes the relative importance of
discrepancies across basis components. For this work, we take
$
W = \operatorname{diag}\!\left( w_\alpha^2(\nu,\sigma) \right),
$
and $g=\|\cdot\|_2$ for consistency with \textbf{Assumption}~\ref{ass: A1}.

We next determine the conformal threshold.
Let $\hat{q}$ denote the
$
\left\lceil (d_0 + 1)(1 - \gamma)\right\rceil/d_0 $
quantile of the empirical distribution of the calibration scores
$\{s_i\}_{i=1}^{d_0}$. This threshold determines the admissible level of deviation
between predicted and observed coefficients. For a new input distribution $X_0$, we first construct a prediction set in the
coefficient space,
\begin{align*}
\mathcal{E}(X_0)
= \left\{ a \in \mathbb{R}^{|T_R|} :
g\Big(
W^{1/2} \big( \theta(\hat{f}(X_0)) - a \big)
\Big)
\le \hat{q}
\right\}.
\end{align*}
Finally, we map the coefficient-level set back to distribution space and define the CP set as
\begin{align*}
\mathcal{Q}(X_0)
= \Big\{
y_a(z) = \textstyle\sum_{\alpha \in T_R} a_\alpha \beta_\alpha(z)
:
a \in \mathcal{E}(X_0)
\Big\}.
\end{align*}

\textbf{Discussion of \textbf{Assumption} \ref{ass: A1}. }This assumption specifies that each output distribution $ Y \in \mathcal{Y} $ admits a representation whose projection coefficients $ \theta(Y)=\{\theta_\alpha(Y)\}_{\alpha\in\mathbb{Z}^l} $ have finite energy in an anisotropic Sobolev sense. Concretely, this requires the weighted sum $ \sum_{\alpha\in\mathbb{Z}^l} \theta_\alpha^2\, w_\alpha^2(\nu,\sigma) $ to be bounded, which enforces decay of high-frequency components at rates that may differ across coordinates. This condition effectively restricts $ Y $ to be sufficiently smooth, while allowing heterogeneous regularity across dimensions. Hence, imposing this assumption amounts to requiring only that the output distributions exhibit reasonable, possibly direction-dependent smoothness, an assumption that is both mild and well aligned with real-world distributional settings.
\subsection{Theoretical Properties}
\textbf{Coverage Guarantee. }We show that the coverage guarantee in
\cref{eq: CovGuaren} holds for both the coefficient-level prediction set
$\mathcal{E}$ and the induced distribution-level prediction set $\mathcal{Q}$.
This result confirms that the proposed construction satisfies the fundamental
validity requirement of CP.

\begin{theorem} \label{thm: MarginCov}
For any distribution $X_0 \in \mathcal{X}$, let $Y_0$ denote the corresponding output distribution, with approximation density $\tilde y_0$. Under the exchangeability of the data, the following coverage guarantees
hold for any $\gamma\in (0,1)$:
\begin{align*}
1 - \gamma
\;\le\;
&\mathbb{P}\!\left( \tilde y_0 \in \mathcal{Q}(X_0) \right)
\;\le\;
1 - \gamma + \frac{\bar{d}}{1 + d_0}, 
\quad 1 - \gamma
\;\le\;\mathbb{P}\!\left( \theta(\tilde y_0) \in \mathcal{E}(X_0) \right)
\;\le\;
1 - \gamma + \frac{\bar{d}}{1 + d_0},
\end{align*}
where $\bar{d}$ denotes the maximum number of ties in the set
$
\{s_1, s_2, \ldots, s_{d_0}, s_0\},
 \text{ with }~ s_0 = s(\hat f(X_0), Y_0).
$
\end{theorem}

The proof of \cref{thm: MarginCov} is deferred to Appendix~\ref{proof: MarginCov}. The coverage statements are equivalent:
\begin{align}\label{eq: eqivofEQ}
\tilde y_0 \in \mathcal{Q}(X_0) ~ 
\text{ if and only if } ~
\theta(\tilde y_0) \in \mathcal{E}(X_0),
\end{align}
which follows from the definition of the prediction sets and the
one-to-one correspondence between coefficient vectors and their associated
truncated density representations.

\textbf{Volume Characterization. }We next characterize the volume of the proposed prediction set. Since $\mathcal{E}(X_0)$ is an ellipsoid in coefficient space, its volume has a closed form. By \cref{eq: eqivofEQ}, $\mathcal{Q}(X_0)$ is the image of $\mathcal{E}(X_0)$ under the linear coefficient-to-density map, which is an isometry on the truncated basis span. Hence, $\mathcal{Q}(X_0)$ and $\mathcal{E}(X_0)$ have identical induced volume, yielding the following result.

\begin{theorem}[Volume]\label{thm: volMarg}
Let $t = |T_R|$, and $g \equiv \|\cdot\|_2$. Recall the Gamma function,
$\Gamma(z)=\int_0^\infty t^{z-1}e^{-t}\,dt$.
The volume of the prediction set $\mathcal{Q}(X_0)$ is given by
\begin{align*}
\operatorname{Vol}\big( \mathcal{Q}(X_0) \big)
=
\dfrac{\left(\sqrt{\pi}\hat{q}\right)^t}{\left({\Gamma\!\left( \tfrac{t}{2} + 1 \right)}
\sqrt{\det W}\right)}.
\end{align*}
\end{theorem}
The proof of Theorem~\ref{thm: volMarg} is given in Appendix~\ref{proof of the Prediction Set Volume}. Since $w_\alpha(\nu,\sigma)$ admits a uniform lower bound
by a constant $m$ depending only on $\nu$ and $\sigma$, we obtain the following as an immediate consequence.

\begin{corollary}\label{coro: vol}
Using the same notation as in \cref{thm: volMarg}, for $c\coloneqq {\sqrt{\pi}}/{m}$, we have
\begin{align*}
\operatorname{Vol}\left( \mathcal{Q}(X_0) \right)
=
\mathcal{O}\left(
\big(c\hat q\big)^{t}\big/\Gamma\left( {t}/{2} + 1 \right)
\right).
\end{align*}
\end{corollary}
Corollary \ref{coro: vol} shows that the volume of the CP set depends
solely on the quantile $\hat{q}$ and truncation $R$.

\textbf{Uniform Sampling.}
Since
$\mathcal{E}(X_0)$ is an ellipsoid in the coefficient space, uniform samples can
be generated using standard methods \cite{l2025non}. The one-to-one
correspondence between $\mathcal{E}(X_0)$ and $\mathcal{Q}(X_0)$ allows these
samples to be mapped to the distribution space, yielding uniform
samples from $\mathcal{Q}(X_0)$.

\textbf{Calibration Set Size. }Beyond interpretability and sampling considerations, an additional practical
question in CP concerns the size of the calibration set
required to achieve a desired coverage level. In practice, the calibration size
cannot be chosen arbitrarily large, as the computational cost of conformal
prediction scales linearly with the number of calibration samples. The following
result addresses this issue by characterizing the conditional coverage behavior
induced by a finite calibration set.

\begin{theorem}\label{thm: marg gamma}
Let
$
\ell := \lfloor (d_0 + 1)\gamma \rfloor .
$
Then the conditional coverage probability satisfies
\begin{align*}
\mathbb{P}\left(
\tilde y_0 \in \mathcal{Q}(X_0)
\ \middle|\
\mathcal{C}
\right)
~~\sim~~
\mathrm{Beta}(d_0 + 1 - \ell, \ell).
\end{align*}
\end{theorem}
The proof of \cref{thm: marg gamma} is provided in Appendix~\ref{proof: T4.4}. In practice,
one is often interested in empirical coverage behavior evaluated over a finite
collection of test instances that share the same calibration set. The following result extends
\cref{thm: marg gamma} to this scenario by characterizing the distribution of the
number of covered test points.

\begin{corollary}\label{coro: binombetamargin}
Let
$\{(X_0^{(j)}, Y_0^{(j)})\}_{j=1}^t$ be $t$ exchangeable test pairs, independent of
the calibration sample. Define
$
I := \sum_{j=1}^t \mathbf{1}\left\{\tilde y_0^{(j)} \in \mathcal{Q}(X_0^{(j)}) \right\}, \text{ and } ~ \ell := \lfloor (d_0 + 1)\gamma \rfloor.
$
Then, we have
\begin{align*}
I ~\big|~\mathcal{C}
~~\sim~~
\mathrm{BetaBinomial}\left(t,\, d_0 + 1 - \ell, \ell \right).
\end{align*}
\end{corollary}
The proof of Corollary~\ref{coro: binombetamargin} is provided in
Appendix~\ref{proof: binombetamargin}.

%% file: sections/5_Adaptive_Conformal_for_Conditional_Guarentee.tex
\section{Adaptive CP Framework with Conditional Coverage Guarantees} \label{sec: AdapCP}
While the coverage guarantee in \cref{eq: CovGuaren} provides strong
distribution-free uncertainty control on average, it does not ensure reliable
coverage for a fixed test input. In particular, marginal validity permits the
coverage probability to vary across test instances and may therefore offer
limited guarantees at the individual level. In many applications, however, it is
desirable to provide uncertainty guarantees that adapt to the specific test input
and are thus more informative on an instance-by-instance basis
\cite{guan2019conformal,angelopoulos2021gentle,ding2023class,gibbs2025conformal,lugosi2025conformal}.
Based on this motivation, following the framework of
\cite{thurin2025optimal}, we develop an Adaptive CP procedure that targets conditional coverage. Specifically, our goal is to
guarantee
\begin{align}\label{eq: ConEq}
\mathbb{P}\!\left( \tilde y_0 \in \mathcal{Q}(X_0) \,\middle|\, X_0 \right)
\xrightarrow[]{\mathbb{P}} 1 - \gamma .
\end{align}
Here, 
$\mathbb{P}\left( \tilde y_0 \in \mathcal{Q}(X_0) \mid X_0 \right)$
is with respect to the randomness of the test output $Y_0$ given $X_0$, whereas convergence in probability is with respect to the calibration sample. We first present the construction of the prediction set for Adaptive CP, and then establish its asymptotic conditional coverage guarantee in \cref{eq: ConEq}. We further analyze the dependence of the coverage on
the size of the calibration set. We summarize our method in \cref{alg:adaptive_cp_ddr} in the Appendix.

\subsection{Construction of Adaptive Prediction Set}
Let $D : \mathcal{X} \times \mathcal{X} \to \mathbb{R}_{\ge 0}$ denote a
user-specified distance on the input space $\mathcal{X}$. For a given test distribution $X_0$, we construct a local
calibration subset by selecting the $k$ nearest input distributions from the
calibration set with respect to the distance $D$. Specifically, we
define the index set
$$
\mathcal{I}_{X_0}
\coloneqq
\arg\min_{i \in [d_0]}^{(k)} D(X_0, X_i^c),
$$
and the associated local calibration set
$
\mathcal{C}_{X_0}
\coloneqq
\{ (X_i^c, Y_i^c) : i \in \mathcal{I}_{X_0} \}.
$
As in the global construction, we take
$
W = \operatorname{diag}\!\left( w_\alpha^2(\nu,\sigma) \right), 
\text{ and }
g = \|\cdot\|_2
$
. Restricting attention to the calibration pairs
in $\mathcal{C}_{X_0}$, we compute localized conformal scores. For the
$i^{\text{th}}$ element $(X, Y) \in \mathcal{C}_{X_0}$, we define our score as
\begin{align*}
s_i^{X_0}(X, Y)
\;:=\;
g\Big(
W^{1/2}
\Big(
\theta(\hat f(X)) - \theta(\tilde y)
\Big)
\Big).
\end{align*}
We now define the local conformal threshold
$\hat q_{X_0}$ as the
$\left\lceil (k+1)(1-\gamma) \right\rceil$-th order statistic of the collection
$\{ s_i^{X_0} \}_{i=1}^{|\mathcal{C}_{X_0}|}$. This threshold is then used to
construct a coefficient-level prediction region,
\begin{align*}
\mathcal{E}(X_0)
\coloneqq
\left\{
a \in \mathbb{R}^{|T_R|}
:
g\Big(
W^{1/2}
\big(
\theta(\hat f( X_0)) - a
\big)
\Big)
\le
\hat q_{X_0}
\right\}.
\end{align*}
Finally, we obtain the prediction set by mapping
$\mathcal{E}(X_0)$ back through the basis expansion, yielding
\begin{align*}
\mathcal{Q}(X_0)
\coloneqq
\Big\{
y_a(x)
=
\textstyle\sum_{\alpha \in T_R} a_\alpha \beta_\alpha(x)
:
a \in \mathcal{E}(X_0)
\Big\}.
\end{align*}

\subsection{Theoretical Properties}
\textbf{Coverage Guarantee.} To establish the conditional coverage property stated in \cref{eq: ConEq}, we
impose the following regularity assumption on the conditional distribution of
the nonconformity scores. A discussion of this assumption is provided at the end of
the section.

\begin{assumption}\label{ass: A2}
Recall that the calibration pairs $(X_1^c, Y_1^c), \ldots, (X_{d_0}^c, Y_{d_0}^c)$ and the test pair $(X_0, Y_0)$ are i.i.d.\ from a joint distribution $\mathcal{P}$. Assume that for almost every input distribution $x$, the conditional law $\mathcal{P}_{S \mid X = x}$ of the score
$$
s_0
=
g\Big(
W^{1/2}
\big(
\theta(\hat{f}(X_0)) - \theta(\tilde{y}_0)
\big)
\Big), \text{ given }~ X_0 = x,
$$
is absolutely continuous with respect to the Lebesgue measure on its convex
support $\mathrm{Supp}(\mathcal{P}_{S \mid X=x})$. Moreover, for any $R > 0$,
suppose that the conditional density $p_S(\cdot \mid x)$ satisfies, for all
$$
s \in \mathrm{Supp}(\mathcal{P}_{S \mid X=x})
\cap \mathrm{Ball}_{\|\cdot\|}(R),
$$
the bounds $$
\lambda_R^{x}
 \le
p_S(s \mid x)
 \le
\Delta_R^{x},
$$
where $\lambda_R^{x}$ and $\Delta_R^{x}$ are local lower and upper bounds,
respectively.
\end{assumption}
Under \textbf{Assumptions}~\ref{ass: A2} and~\ref{ass: A1}, we show that the prediction set
$\mathcal{Q}$ achieves the conditional coverage guarantee in
\cref{eq: ConEq}. This guarantee is stated formally in the following theorem.

\begin{theorem}[Asymptotic Conditional Coverage] \label{thm: CondCov}
Let $k$ denote the number of nearest neighbors used to estimate the local
quantile around each test input $X_0$. Under \textbf{Assumptions}
\ref{ass: A1} and \ref{ass: A2}, assume further that
$k \to \infty$ and $k / d_0 \to 0$ as $d_0 \to \infty$, and that
$$
\widehat F_k(s \mid x)
\xrightarrow[]{\mathbb{P}}
F_x(s)
$$
for $\mathbb{P}_X$-almost every $x$ and every continuity point $s$ of $F_x$.
Then, for any $\gamma \in (0,1)$,
\begin{align*}
\mathbb{P}\!\left(
\tilde y_0 \in \mathcal{Q}(X_0)
\mid X_0
\right)
\xrightarrow[]{\mathbb{P}}
1 - \gamma,
\quad
\text{as } d_0 \to \infty.
\end{align*}
\end{theorem}
The proof of \cref{thm: CondCov} is provided in Appendix~\ref{proof: CondCov}.

\begin{remark}[Discussion of the Condition
$\widehat F_k \to F_x$ in Theorem~\ref{thm: CondCov}]
The convergence condition
$\widehat F_k(s\mid x)\xrightarrow{\mathbb P}F_x(s)$
is a standard consistency requirement in nonparametric
estimation of conditional distributions, closely related
to regularity conditions employed in previous studies
of conditional conformal prediction
\cite{lei2014distribution,guan2022localizedconformalpredictiongeneralized}.
Intuitively, as $k\to\infty$, the increasing number of
local calibration samples reduces estimation variance,
while $k/d_0\to 0$ ensures that the selected neighbors
concentrate around the test input. Consequently, the
empirical score distribution approaches the true
conditional distribution, yielding a consistent estimate
of the conditional quantile.

\begin{figure*}[t]
    \centering
    \begin{subfigure}{0.49\textwidth}
        \centering
        \includegraphics[width=\linewidth]{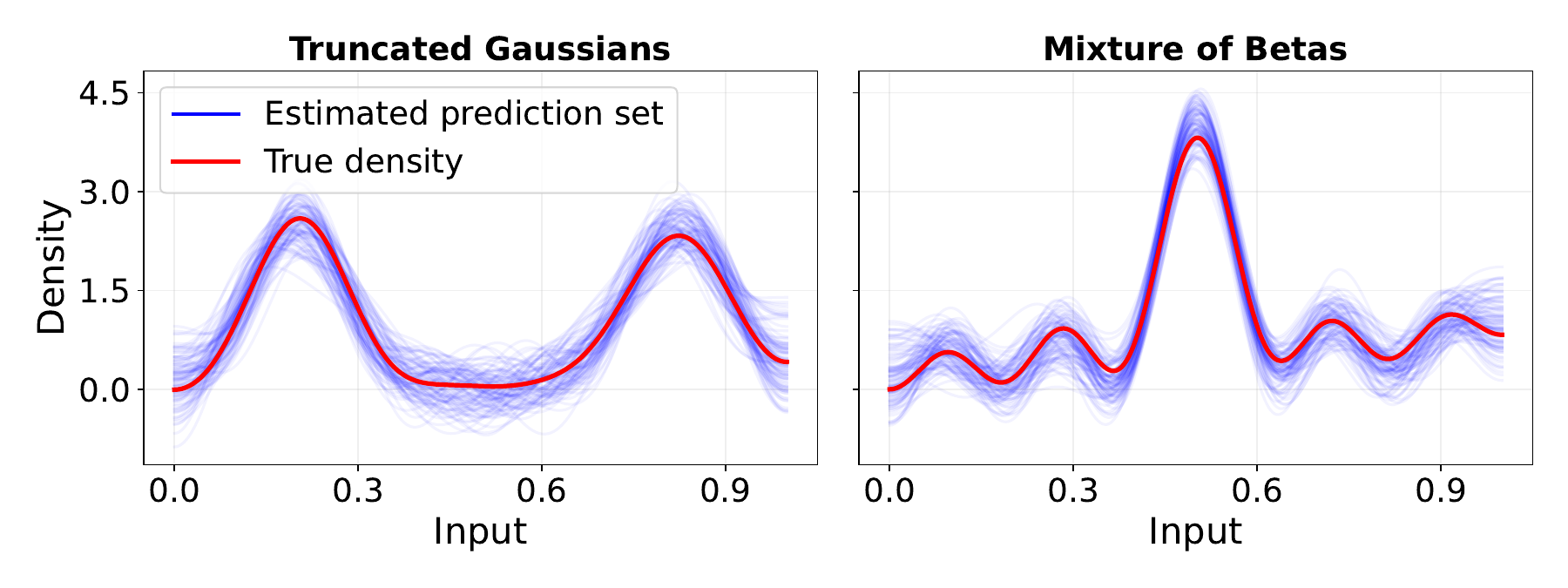}
        \caption{Synthetic Datasets} 
    \end{subfigure}
    \hfill
    \begin{subfigure}{0.49\textwidth}
        \centering
        \includegraphics[width=\linewidth]{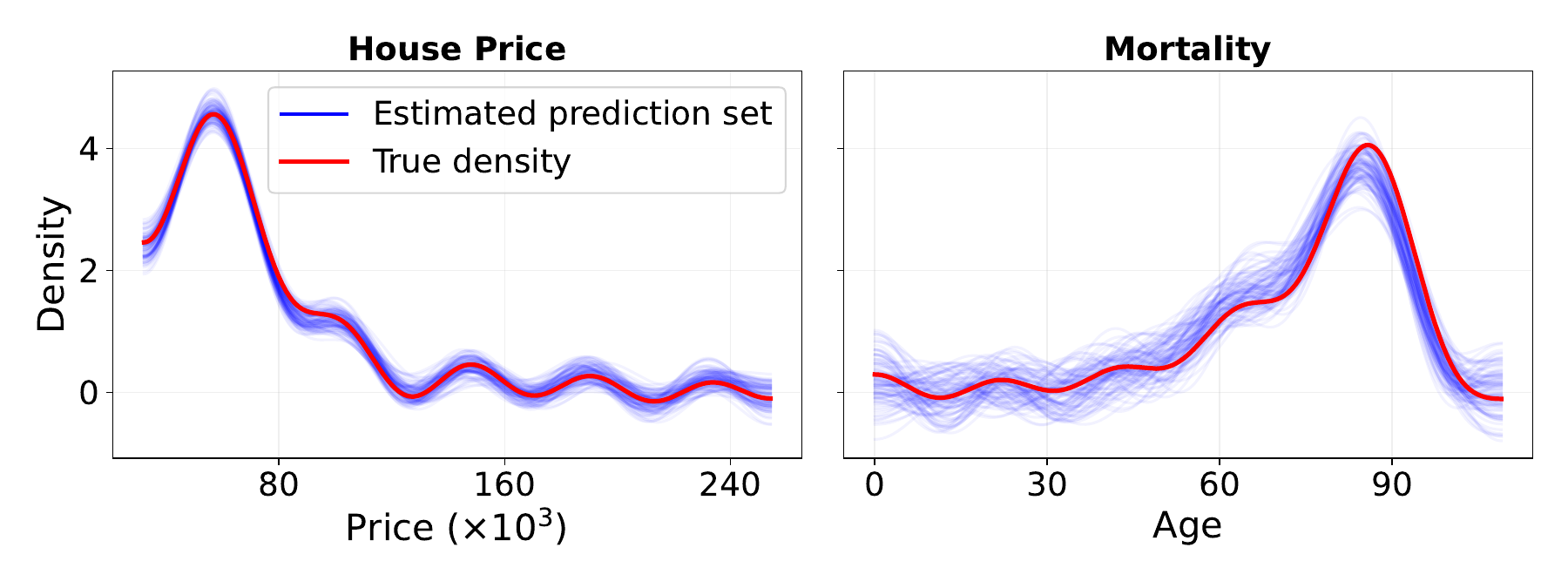}
        \caption{Real Datasets }
    \end{subfigure}
    \caption{
Visualization of distributions uniformly sampled from the proposed conformal prediction sets (blue), compared with the true density (red). Panel (a) presents results on synthetic datasets (Truncated Gaussians and Mixture of Betas), while panel (b) shows results on real-world datasets (House Price and Mortality).
}
    \label{fig: Dist Est}
    \vspace{-10pt}
\end{figure*}
This condition is relatively mild and holds in several
common settings. For instance, it is satisfied when the
conditional score CDF $F_x(s)$ varies continuously with
$x$ under the distance $D$, and the $k$-nearest-neighbor
radius converges to zero. This includes smooth
finite-dimensional parametric families of input
distributions, such as Gaussian families, with
continuously varying conditional score distributions.
\end{remark}

\textbf{Volume Characterization and Uniform Sampling. }As in the marginal setting, the proposed conditional prediction sets inherit
favorable geometric properties. In particular, since $\mathcal{E}(X_0)$ is an
ellipsoid in the coefficient space, its volume admits a closed-form expression,
and uniform sampling can be performed efficiently using standard methods.

\textbf{Calibration Set Size.}
We next study the effect of the local calibration set size on conditional coverage. We first introduce the following mild assumption.
\begin{assumption}\label{ass: A3}
Fix a test input $X_0$. Assume that the test score
$
S_0 := s(\hat f(X_0), Y_0)
$
and the local calibration scores
$
\{ S_i := s(\hat f(X_i^c), Y_i^c) : i \in \mathcal{I}_{X_0} \}
$
are i.i.d. according to a distribution with
continuous cumulative distribution function $F_{X_0}$.
\end{assumption}

Under this assumption, we can explicitly characterize how the achieved
conditional coverage depends on the size of the local calibration set through the following result.

\begin{theorem}\label{thm: local_gamma}
Fix $\gamma \in (0,1)$ and let $\ell = \lfloor (k+1)\gamma \rfloor$.
Under \textbf{Assumptions} \ref{ass: A1} and \ref{ass: A3}, the conditional
coverage probability satisfies
\begin{align*}
\mathbb{P}\!\left(
\tilde y_0 \in \mathcal{Q}(X_0)
\mid \mathcal{C}_{X_0}, X_0
\right)
\sim \mathrm{Beta}(k+1-\ell,\,\ell).
\end{align*}
\end{theorem}
The proof of \cref{thm: local_gamma} is provided in
Appendix~\ref{proof: localgamma}.

\textbf{Discussion of \textbf{Assumption} \ref{ass: A2}. }This assumption requires the nonconformity score distribution to have a well-behaved density, so that empirical quantiles from nearby calibration samples remain stable. It rules out pathological cases where small perturbations of the input distribution cause abrupt changes in the score distribution. The bounded-density condition captures local smoothness of the uncertainty structure around each test input, as is standard in conditional and local CP analyses.

\textbf{Discussion of \textbf{Assumption} \ref{ass: A3}. } This assumption requires the test score and the local calibration scores to be i.i.d. with a continuous distribution; it ensures exchangeability within the local neighborhood and rules out ties among scores. This condition is typically met when the score function is continuous, and the calibration points are drawn from a sufficiently small neighborhood around the test input, so the local score distributions are approximately identical.

%% file: sections/6_Experiment.tex
\section{Experimental Results}\label{sec: Exp}
This section provides numerical validation showing that (i) the proposed framework satisfies the coverage guarantees established in \cref{thm: MarginCov}, while achieving a reasonable trade-off between coverage and prediction set volume; (ii) the number of calibration samples needed to achieve a prescribed coverage level for a given number of test instances follows Corollary~\ref{coro: binombetamargin}; (iii) Adaptive CP conforms to the asymptotic conditional coverage behavior characterized in \cref{thm: CondCov}; and (iv) distributions obtained from the resulting prediction sets closely approximate the true distributions.

\begin{figure*}[t]
    \centering
    \includegraphics[width=0.99\linewidth]{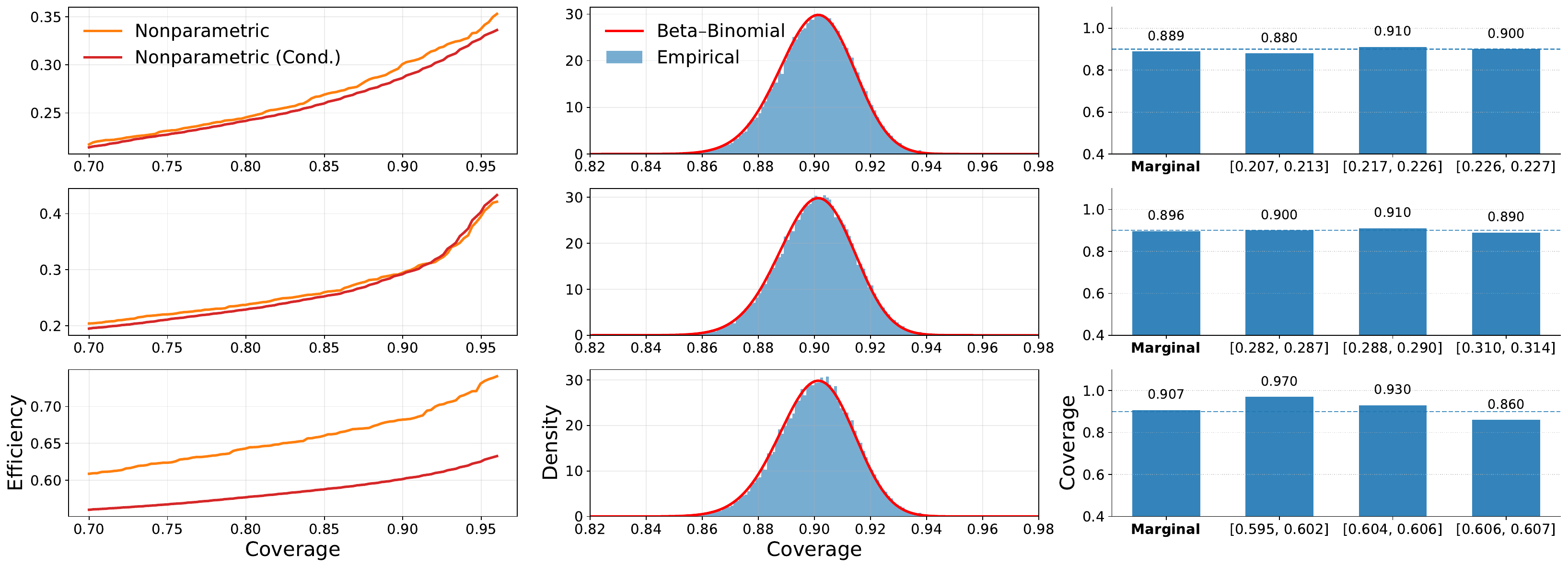}
    \caption{
Results on synthetic datasets using the NDR baseline. Rows correspond to Truncated Gaussian, Mixture of Betas, and 3D Dark Matter datasets (top to bottom). Columns correspond to tasks \textbf{(I)}--\textbf{(III)} (left to right): efficiency-coverage trade-off between adaptive and non-adaptive CP, comparison of empirical coverage with the theoretical Beta-Binomial distribution, and empirical conditional coverage across different input groups, respectively. Dashed horizontal lines in the rightmost column indicate the nominal coverage level.
}
    \label{fig:SyntheticData}
    \vspace{-8pt}
\end{figure*}

\textbf{Overall Setup.}  We evaluate our proposed method on both synthetic and real-world datasets. For synthetic data, we consider one-dimensional mixtures of two truncated Gaussians \cite{oliva2013distribution} and mixtures of three Beta distributions \cite{okano2023distribution}, as well as multidimensional dark-matter datasets in 3D and 6D generated following \cite{klein2014scalable}. For real-world evaluation, we use the house-price dataset studied in \cite{chen2023wasserstein}. As base predictors $\hat f$, we consider three models: nonparametric distribution regression (NDR) \cite{oliva2013distribution}, optimal transport map regression (OTR) \cite{okano2023distribution}, and Wasserstein regression (WR) \cite{chen2023wasserstein}. 

\textbf{Ablation Study. }To further evaluate the stability and practical behavior of our method, we conduct ablation studies on the truncation parameter $R$, inference-time efficiency, and hyperparameter sensitivity, and evaluate performance in the low-data regime using the real-world Mortality dataset from \cite{ghodrati2022distribution}. The corresponding results are reported in Appendices~\ref{app: AblationTrun}, \ref{app: AblationInftime}, \ref{app:  abonHyperparam}, and \ref{app: lowdata}, respectively.
\subsection{Simulation Study}\label{sec: SimStudy}
We evaluate three tasks:
\textbf{(I)} examining the trade-off between the target coverage level $1-\gamma$ and prediction set volume; since the volume depends on the calibration threshold $\hat q$, we instead report results in terms of average $\hat q$ \textit{(Efficiency)} versus $1-\gamma$ to avoid numerical instability; \textbf{(II)} validating the relationship between calibration-set size and empirical coverage as characterized in Corollary~\ref{coro: binombetamargin}; and \textbf{(III)} for adaptive CP, empirically verifying the conditional-coverage guarantee in \cref{thm: CondCov} using the evaluation protocol in Appendix~\ref{app: ConfMeasure}. Full experimental details are provided in Appendix~\ref{App: Símtudy}.

\textbf{Truncated Gaussians and Mixture of Betas. }
Results for NDR tasks \textbf{(I)}--\textbf{(III)} under both truncated Gaussian and mixture-of-Betas settings are shown in \cref{fig:SyntheticData}; OTR and WR results are shown in \cref{fig:GaussBetaOTR} and \cref{fig:GaussBetaWR}, respectively. For \textbf{(I)}, across all models and both settings, the adaptive variant consistently produces prediction sets with comparable or smaller volumes than its non-adaptive counterpart, demonstrating the effectiveness of adaptivity in reducing prediction set size. Both variants exhibit a clear efficiency-coverage trade-off that is consistent with trends reported in the existing literature \cite{zargarbashi2023conformal, zhang2024conformal}. For \textbf{(II)}, the results closely match the Beta-Binomial relationships derived in Corollary~\ref{coro: binombetamargin}, demonstrating accurate calibration across varying calibration sample sizes. For \textbf{(III)}, adaptive CP closely tracks the target conditional coverage, with deviations bounded by 0.04 (Truncated Gaussians) and 0.06 (Mixtures of Betas).


\textbf{Dark Matter Data.}
Due to methodological limitations, only NDR is applicable to the multidimensional setting. Results for tasks \textbf{(I)}--\textbf{(III)} are presented in \cref{fig:SyntheticData} for 3D data and \cref{fig:DM6D} for 6D data. Consistent with the one-dimensional experiments, task \textbf{(I)} demonstrates that adaptive CP achieves substantially better efficiency than its non-adaptive counterpart, while both methods exhibit the expected efficiency-coverage trade-off. For task \textbf{(II)}, the empirical results in both dimensions closely match the theoretical Beta-Binomial distribution established in Corollary~\ref{coro: binombetamargin}. Finally, for task \textbf{(III)}, adaptive CP achieves near-nominal conditional coverage, with maximum deviations of 0.04 and 0.02 for the 3D and 6D datasets, respectively. These findings further support the effectiveness of our adaptive framework in higher-dimensional distributional settings.

\textbf{Visualization. }Figure~\ref{fig: Dist Est}(a) demonstrates the effectiveness of the proposed prediction sets via uniform sampling from each set and overlaying with the true density across datasets.
\subsection{House Price Dataset}
\begin{figure}[t]
    \centering
    \includegraphics[width=0.99\linewidth]{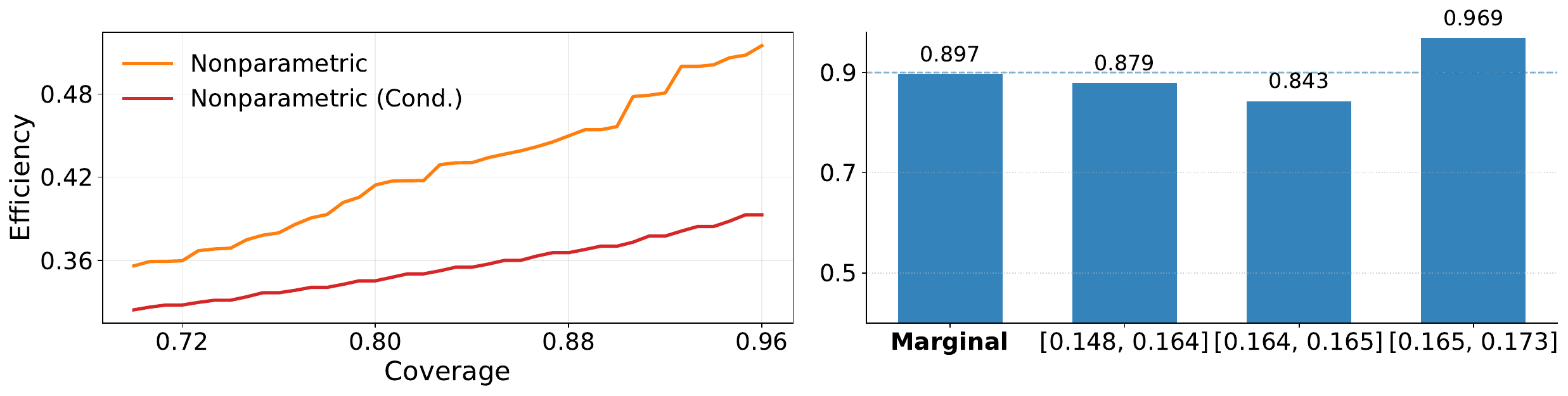}
    \caption{
House Price experiments using the NDR base model. The left panel (task \textbf{(I)}) compares the efficiency-coverage trade-off between adaptive and non-adaptive CP, while the right panel (task \textbf{(III)}) reports empirical conditional coverage across different input groups, with the dashed line indicating the nominal coverage level.
}
    \label{fig:HousePrice main}
    \vspace{-15pt}
\end{figure}

For the real-world evaluation, we consider the House Price dataset introduced in \cite{chen2023wasserstein}, with further details provided in Appendix~\ref{app: Houseprice}. We focus on tasks \textbf{(I)} and \textbf{(III)} defined in \cref{sec: SimStudy}, omitting task \textbf{(II)} due to the limited sample size. Results for NDR are presented in \cref{fig:HousePrice main}, while the corresponding results for OTR and WR are reported in \cref{fig:HousePriceOTRWR}. Consistent with the simulation study, task \textbf{(I)} demonstrates that both adaptive and non-adaptive CP exhibit the expected efficiency-coverage trade-off reported in the literature \cite{zargarbashi2023conformal}, with the adaptive variant consistently achieving comparable or smaller prediction set volumes across all models. For task \textbf{(III)}, adaptive CP achieves near-nominal conditional coverage, with deviations bounded by 0.06. These findings further support the effectiveness and reliability of our adaptive framework in real-world distributional regression settings.

\textbf{Visualization. }Figure~\ref{fig: Dist Est}(b) visualizes the prediction sets by uniformly sampling distributions from them and overlaying these samples with the true density.
\section{Conclusion}\label{sec: Conc}
In this paper, we study conformal prediction (CP) for Distribution-to-Distribution Regression (DDR), where directly applying classical Split CP yields prediction sets that lack interpretability despite preserving marginal coverage guarantees. To address this limitation, we propose a CP framework based on projection onto a truncated orthonormal basis, enabling explicit volume characterization and efficient uniform sampling while maintaining coverage validity. We further develop an adaptive variant that incorporates local calibration and achieves asymptotic conditional coverage under mild assumptions. Extensive experiments on synthetic and real-world datasets, spanning both low- and high-dimensional settings and multiple DDR models, validate our theoretical results and demonstrate improved prediction efficiency through adaptivity. Despite these promising results, our framework does not directly accommodate covariate shift~\cite{tibshirani2019conformal}, and extending it to non-exchangeable settings, together with improving local calibration in high-dimensional distribution spaces, represents an interesting direction for future research.

\newpage
\begin{ack}
This paper is dedicated to Professor Richard G. Baraniuk on the occasion of his 60th birthday.

This research is supported by the National Research Foundation Singapore under the AI Singapore Programme (AISG Award No: AISG2-TC-2023-012-SGIL). This research is also supported by the Ministry of Education, Singapore, under the Academic Research Fund Tier 1 (FY2023) (A-8002040-00-00, A-8002039-00-00), the NUS Presidential Young Professorship Award (A-0009807-01-00), the NUS Artificial Intelligence Institute--Seed Funding (A-8003062-00-00), and the Cross Faculty Grant 2025, CFG25-012 (A-8004460-00-00).
\end{ack}

%% file: sections/appendix1.tex
\section*{Table of Notation}

\begin{table}[h]
\renewcommand*{\arraystretch}{1.3}
    \begin{tabularx}{\textwidth}{p{0.4\textwidth}X}
    \toprule
    $\mathbb{R}^k$ & $k$-dimensional Euclidean space\\
    $\|\cdot\|_2$ & Euclidean norm \\
    $\text{Ball}_{\|\cdot\|}(R)$ & a ball of radius $R$, induced by $\|\cdot\|$\\
    $\Gamma(\cdot)$ & Gamma function\\
    $\mathbf{1}\{\cdot\}$ & Indicator function\\
    $\text{Beta}(\cdot,\cdot)$ &  Beta distribution \\
    $\text{BetaBinomial}(\cdot,\cdot,\cdot)$ &  Beta-Binomial distribution \\
    $\theta(Y)$ &  collection of projection
coefficients associated with distribution $Y$\\
$\theta_{\alpha}(Y)$ &  expansion coefficients of $Y$ under basis $\beta_{\alpha}$\\
    $L_2(\Omega)$ & space of square-integrable functions on $\Omega$ with respect to the Lebesgue measure.\\
    $\mathcal{P}(X)$ & powerset of a set $X$\\
    $\text{Vol}(X)$ & volume of a set $X$\\
    $|X|$ & cardinality of set $X$\\
    $\mathcal{X}$ & class of input distributions\\
    $\mathcal{Y}$ & class of output distributions\\
    $\mathcal{P}$ & probability measure defined on the class of input distribution\\
    $\mathcal{C}$ & calibration set\\
    $\mathcal{E}$ &  coefficient-level prediction set\\
    $\mathcal{Q}$ &  distribution-level prediction set\\
    $R$ & truncation parameter\\
    $\gamma$ & target miscoverage level \\
    $d_t$ & number of Training data points \\
    $d_0$ & number Calibration data points\\
    $m_i$ & number of samples for the i$^{\text{th}}$ input distribution \\
    $n_i$ & number of samples for the i$^{\text{th}}$ output distribution \\
    $\hat{f}$ & prediction model\\
    $\{s_i\}_{i=1}^{d_0}$ & nonconformity score\\
    $\hat{q}$ & the $ \lceil (d_0+1)(1-\gamma)\rceil $-th order statistic of $\{s_i\}_{i=1}^{d_0}$\\
    $\mathcal{I}_X$ & set of $k$ nearest neighbor to $X$ in the calibration set.\\

    \toprule
     \end{tabularx}
\end{table}

\clearpage
\begin{center}
{\bf \Large{Appendix of "Conformal Prediction for Distribution-to-Distribution Regression"}}
\end{center}

\DoToC

\section{Background on Conformal Prediction}\label{app: backgroundConf}
\subsection{Introduction}
\textbf{Motivation. }Deep learning and modern AI have dramatically improved performance across a wide range of tasks, from natural language processing and computer vision to reinforcement learning and beyond, often producing impressively accurate predictions on complex data. However, these methods are fundamentally probabilistic and data-driven, meaning their outputs are point estimates with associated uncertainties rather than guaranteed truths. In high-risk settings such as medical diagnosis, autonomous driving, or financial risk assessment, relying solely on a single prediction can be dangerous: an overconfident yet incorrect prediction can lead to misdiagnosis, unsafe decisions, or substantial economic loss. Traditional model confidence scores are often poorly calibrated and can be misleading, especially when the model encounters rare or out-of-distribution cases. 

Conformal Prediction offers a principled solution to this challenge by providing statistically valid prediction sets or intervals that are guaranteed to contain the true outcome with a user-specified probability level, such as 90 \% or 95\%. At a high level, conformal methods work by measuring how “non-conforming” a new input looks relative to past data and then using this information to construct uncertainty sets that satisfy finite-sample coverage guarantees under minimal assumptions (essentially just exchangeability of the data). This makes the Conformal Prediction model-agnostic and distribution-free, meaning it can be applied on top of any predictive model, including neural networks, without assuming a specific data distribution, and still deliver reliable uncertainty quantification essential for trustworthy decision-making in safety-critical applications.

\textbf{Full (Transductive) Conformal Prediction \cite{shafer2008tutorial,angelopoulos2021gentle}. }
Historically, full Conformal Prediction is the original form of Conformal Prediction, introduced before its more computationally efficient variants such as Split Conformal Prediction. In the full conformal framework, one considers a collection of $d+1$ exchangeable data points ${(X_ 1, Y_1),\ldots,(X _d, Y_d),(X_{d+1}, Y_{d+1})}$, where the learner observes ${(X_i, Y_i)}_{i=1}^d$ and the test input $X_{d+1}$, and aims to construct a prediction set for the unknown response $Y_{d+1}$. Unlike Split Conformal Prediction, where the training and calibration stages are separated, full Conformal Prediction allows the predictive model to be trained on all available data, including the candidate test label.
The core idea is to loop over candidate outputs $y\in\mathcal{Y}$ and, for each $y$, fit a model $\hat f^{\,y}$ on the augmented sample
\begin{align}
\{(X_1,Y_1),\ldots,(X_d,Y_d),(X_{d+1},y)\}.
\end{align}
Crucially, the learning algorithm used to produce $\hat f^{\,y}$ must be permutation-invariant (symmetric) in its training data.
\begin{algorithm}[t]
\caption{Conformal Prediction for Distribution-to-Distribution Regression}
\label{alg:cp_ddr}
\begin{algorithmic}[1]

\Require Calibration set $\mathcal{C}=\{(X_i^c,Y_i^c)\}_{i=1}^{d_0}$ with $Y_i^c=\{N_{ij}^c\}_{j=1}^{n_i^c}$,
test input $X_0$,
truncation parameter $R$,
metric $g$,
weight matrix $W$,
miscoverage level $\gamma\in(0,1)$,
and parameters $(\nu,\sigma,\bar{A})$ satisfying Assumption~\ref{ass: A1}.

\Ensure Prediction set $\mathcal{Q}(X_0)$.

\vspace{0.3em}
\State Fix an orthonormal basis $\{\beta_i\}_{i\in\mathbb{Z}}$ of $L_2(\Lambda)$.
\State Define the product basis $\beta_\alpha(z)\leftarrow \prod_{r=1}^{l}\beta_{\alpha_r}(z_r)$ for $\alpha\in\mathbb{Z}^l$, $z\in\Lambda^l$.
\State Compute weights $w_\alpha^2(\nu,\sigma)\leftarrow \sum_{r=1}^{l}(\nu_r|\alpha_r|)^{2\sigma_r}$.

\vspace{0.3em}
\State \textbf{Calibration scores:}
\For{$i=1,\ldots,d_0$}
    \State Compute empirical coefficients $\theta_\alpha(\widehat{Y}_i^c)\leftarrow \frac{1}{n_i^c}\sum_{j=1}^{n_i^c}\beta_\alpha(N_{ij}^c)$ for all $\alpha\in T_R$.
    \State Form the truncated density estimate $\tilde{y}_i^c(z)\leftarrow \sum_{\alpha\in T_R}\theta_\alpha(\widehat{Y}_i^c)\,\beta_\alpha(z)$.
    \State Compute score $s_i \leftarrow g\!\left(W^{1/2}\big(\theta(\hat{f}(X_i^c)) - \theta(\tilde{y}_i^c)\big)\right)$.
\EndFor

\vspace{0.3em}
\State \textbf{Calibrate threshold.} Let $\hat{q}$ be the $\left\lceil (d_0+1)(1-\gamma)\right\rceil$-th order statistic of $\{s_i\}_{i=1}^{d_0}$.

\vspace{0.3em}
\State \textbf{Prediction set in coefficient space:} Define 
$$
\mathcal{E}(X_0)\leftarrow\{a\in\mathbb{R}^{|T_R|}: g(W^{1/2}(\theta(\hat f(X_0))-a))\le \hat q\}.
$$

\vspace{0.3em}
\State \textbf{Map back to distribution space:} Define 
$$
\mathcal{Q}(X_0)
\leftarrow \left\{
y_a(z) = \sum_{\alpha \in T_R} a_\alpha \beta_\alpha(z)
\middle|
a \in \mathcal{E}(X_0)
\right\}.
$$

\vspace{0.3em}
\State \textbf{Return} $\mathcal{Q}(X_0)$.

\end{algorithmic}
\end{algorithm}

\begin{algorithm}[t]
\caption{Adaptive Conformal Prediction for Distribution-to-Distribution Regression}
\label{alg:adaptive_cp_ddr}
\begin{algorithmic}[1]

\Require Calibration set $\mathcal{C}=\{(X_i^c,Y_i^c)\}_{i=1}^{d_0}$,
test input $X_0$,
truncation parameter $R$,
distance $\mathcal D$,
number of neighbors $k$,
metric $g$,
weight matrix $W$,
miscoverage level $\gamma$,
and parameters $(\nu,\sigma,\bar{A})$ satisfying Assumption~\ref{ass: A1}.

\Ensure Adaptive prediction set $\mathcal{Q}(X_0)$.

\vspace{0.5em}
\State Fix an orthonormal basis $\{\beta_i\}_{i\in\mathbb{Z}}$ of $L_2(\Lambda)$.
\State Define the product basis $\beta_\alpha(z)\leftarrow \prod_{r=1}^{l}\beta_{\alpha_r}(z_r)$ for $\alpha\in\mathbb{Z}^l$, $z\in\Lambda^l$.
\State Compute weights $w_\alpha^2(\nu,\sigma)\leftarrow \sum_{r=1}^{l}(\nu_r|\alpha_r|)^{2\sigma_r}$.

\vspace{0.5em}
\State \textbf{Select local calibration subset:}
\State Define the index set $
\mathcal{I}_{X_0}
\leftarrow
\arg\min_{i \in [d_0]}^{(k)} D(X_0, X_i^c),
$
\State Define local calibration set $\mathcal{C}_{X_0}\leftarrow\{(X_i^c,Y_i^c): i\in I_{X_0}\}$.

\vspace{0.5em}
\State \textbf{Compute localized calibration scores:}
\For{$i\in I_{X_0}$}
    \State Compute empirical coefficients $\theta_\alpha(\widehat{Y}_i^c)\leftarrow \frac{1}{n_i^c}\sum_{j=1}^{n_i^c}\beta_\alpha(N_{ij}^c)$ for all $\alpha\in T_R$.
    \State Form the truncated density estimate $\tilde{y}_i^c(z)\leftarrow \sum_{\alpha\in T_R}\theta_\alpha(\widehat{Y}_i^c)\,\beta_\alpha(z)$.
    \State Compute score $s_i^{X_0} \leftarrow g\!\left(W^{1/2}\big(\theta(\hat{f}(X_i^c)) - \theta(\tilde{y}_i^c)\big)\right)$.
\EndFor

\vspace{0.5em}
\State \textbf{Calibrate local threshold:} Let $\hat q_{X_0}$ be the $\lceil (k+1)(1-\gamma)\rceil$-th order statistic of
$\{s_i^{X_0}\}_{i\in I_{X_0}}$.

\vspace{0.5em}
\State \textbf{Prediction set in coefficient space:} Define 
$$
\mathcal{E}(X_0)\leftarrow\{a\in\mathbb{R}^{|T_R|}: g(W^{1/2}(\theta(\hat f(X_0))-a))\le \hat q_{X_0}\}.
$$

\vspace{0.3em}
\State \textbf{Map back to distribution space:} Define 
$$
\mathcal{Q}(X_0)
\leftarrow \left\{
y_a(z) = \sum_{\alpha \in T_R} a_\alpha \beta_\alpha(z)
\middle|
a \in \mathcal{E}(X_0)
\right\}.
$$

\vspace{0.3em}
\State \textbf{Return} $\mathcal{Q}(X_0)$.

\end{algorithmic}
\end{algorithm}
Given a nonconformity score function $s(\cdot,\cdot;\cdot)$, we compute the scores
\begin{align}
s_i^{\,y} := s\!\bigl(X_i,Y_i;\hat f^{\,y}\bigr),\quad i=1,\ldots,d,
\qquad
s_{d+1}^{\,y} := s\!\bigl(X_{d+1},y;\hat f^{\,y}\bigr).
\end{align}
Let $\widehat q^{\,y}$ be the empirical conformal threshold defined as the
$\left\lceil (d+1)(1-\alpha)\right\rceil$-th order statistic of
$\{s_1^{\,y},\ldots,s_{d+1}^{\,y}\}$ (equivalently, the empirical quantile at level
$\left\lceil (d+1)(1-\alpha)\right\rceil/(d+1)$), i.e.,
\begin{align}
\widehat q^{\,y}
:=
\operatorname{Quantile}\!\left(
s_1^{\,y},\ldots,s_{d+1}^{\,y};
\frac{\left\lceil (d+1)(1-\alpha)\right\rceil}{d+1}
\right).
\end{align}
We then include all candidates whose test score is no larger than this threshold:
\begin{align}
\mathcal{C}(X_{d+1})
:=
\Bigl\{\, y\in\mathcal{Y} \;:\; s_{d+1}^{\,y}\le \widehat q^{\,y} \Bigr\}.
\end{align}
Intuitively, $\mathcal{C}(X_{d+1})$ collects exactly those $y$ for which the augmented point
$(X_{d+1},y)$ looks sufficiently ``typical'' relative to the $d$ observed samples under the
model refit on the augmented dataset.

\begin{theorem}[Full conformal coverage guarantee]
\label{thm:full_cp}
Assume $\{(X_i,Y_i)\}_{i=1}^{d+1}$ are exchangeable (in particular, i.i.d.\ is sufficient),
and the learning algorithm producing $\hat f^{\,y}$ is symmetric (permutation-invariant) in its input sample.
Then the full Conformal Prediction set $\mathcal{C}(X_{d+1})$ defined above satisfies the finite-sample marginal coverage guarantee
\begin{align}
\mathbb{P}\!\left( Y_{d+1}\in \mathcal{C}(X_{d+1}) \right)\ge 1-\alpha.
\end{align}
\end{theorem}
While full Conformal Prediction provides strong theoretical guarantees and serves as the conceptual foundation of conformal inference, its requirement to retrain the model for each candidate $y$ makes it computationally expensive in practice. The split conformal procedure considered in this work can be viewed as a scalable special case of full Conformal Prediction, designed to retain validity while significantly reducing computational cost.

\textbf{Split (Inductive) Conformal Prediction. }In \cref{sec: Prelim}, we briefly introduced the framework of split Conformal Prediction. We now revisit this method in greater detail, clarifying its connection to full (transductive) Conformal Prediction.

Split Conformal Prediction can be viewed as a computationally efficient \emph{special case} of full Conformal Prediction.
Recall that the full Conformal Prediction conceptually evaluates all candidate labels $y\in\mathcal{Y}$ by retraining the model on the augmented dataset that includes $(X_{d+1},y)$.
Split Conformal Prediction avoids this repeated refitting by fixing the predictive model in advance and using a held-out calibration set to assess conformity.
This restriction dramatically reduces computational cost while preserving finite-sample validity under exchangeability.

Formally, let $\{(X_i,Y_i)\}_{i=1}^{d}$ be exchangeable samples from an unknown distribution on $\mathcal{X}\times\mathcal{Y}$.
Given a target miscoverage level $\gamma\in(0,1)$, the goal of Conformal Prediction is to construct a set-valued predictor
$\mathcal{C}:\mathcal{X}\to\mathcal{P}(\mathcal{Y})$ such that
\begin{equation}
\mathbb{P}\!\left( Y\in \mathcal{C}(X) \right)\ge 1-\gamma,
\label{eq:CP_goal}
\end{equation}
where the probability is taken over a fresh test pair $(X, Y)$ drawn from the same distribution.

In the split conformal framework, the dataset is randomly partitioned into two disjoint subsets:
a training set and an independent calibration set.
Let $d_0$ denote the size of the calibration set, and write
\begin{align}
\mathcal{I}_{\mathrm{cal}}=\{1,\ldots,d_0\}
\end{align}
for the indices of the calibration points.
A prediction model $\hat f:\mathcal{X}\to\mathcal{Y}$ is trained \emph{only} on the training set and remains fixed throughout the inference procedure.

Given a user-chosen nonconformity score function
\begin{align}
s:\mathcal{Y}\times\mathcal{Y}\to\mathbb{R},
\end{align}
which quantifies the discrepancy between a prediction and an observed response, we compute the calibration scores
\begin{equation}
s_i := s\!\bigl(\hat f(X_i),Y_i\bigr),
\qquad i\in\mathcal{I}_{\mathrm{cal}}.
\label{eq:cal_scores}
\end{equation}
The choice of $s$ is problem-dependent and should reflect the relevant notion of uncertainty for the task at hand. An empirical threshold is then defined as the quantile
\begin{equation}
\hat q
:=
\operatorname{Quantile}\!\left(
\{s_i\}_{i=1}^{d_0};
\frac{\lceil (d_0+1)(1-\gamma)\rceil}{d_0}
\right),
\label{eq:split_quantile}
\end{equation}
equivalently, the $\lceil (d_0+1)(1-\gamma)\rceil$-th order statistic of the calibration scores.
The Split Conformal Prediction set is finally given by
\begin{equation}
\mathcal{C}(x)
:=
\bigl\{ y\in\mathcal{Y} : s\!\bigl(\hat f(x),y\bigr)\le \hat q \bigr\}.
\label{eq:split_set}
\end{equation}

\begin{theorem}[Split conformal coverage guarantee]
\label{thm:split_cp}
Assume $\{(X_i,Y_i)\}_{i=1}^{d}$ are exchangeable.
Then the Split Conformal Prediction set $\mathcal{C}$ defined in \eqref{eq:split_set} satisfies the marginal coverage guarantee
\begin{align}
\mathbb{P}\!\left( Y\in \mathcal{C}(X) \right)\ge 1-\gamma,
\end{align}
for any underlying data distribution and any choice of nonconformity score $s$.
\end{theorem}

\begin{remark}
Split Conformal Prediction arises as a restriction of full Conformal Prediction in which
(i) the predictive model is trained once and held fixed,
and (ii) conformity is evaluated only on an independent calibration subset rather than on the full augmented sample.
This simplification breaks the exact symmetry exploited in full Conformal Prediction but preserves exchangeability between the test point and the calibration scores, which is sufficient to guarantee marginal coverage.
The resulting method trades some statistical efficiency for substantial computational savings, making Split Conformal Prediction the method of choice in large-scale and high-dimensional applications.
\end{remark}
\begin{remark}
Although full Conformal Prediction provides exact finite-sample coverage guarantees, its practical applicability is severely limited for complex or continuous output spaces. In particular, it requires iterating over all candidate outputs $y\in\mathcal{Y}$ and refitting the model on the augmented dataset $(X_{d+1},y)$ for each candidate. When $\mathcal{Y}$ is continuous, high-dimensional, or structured, as in distribution-to-distribution regression, this procedure becomes computationally intractable due to the uncountable output space and repeated model fitting. While exact computation is possible in certain low-dimensional parametric settings with special score structures, such assumptions do not generally hold here. Consequently, full Conformal Prediction is unsuitable for our setting, motivating the use of split (inductive) Conformal Prediction, which preserves distribution-free validity while remaining computationally feasible.
\end{remark}

\textbf{Extensions of Conformal Prediction. }Recent work has extended Conformal Prediction well beyond its classical setting, enabling valid uncertainty quantification in increasingly complex and modern learning problems.

\textit{Group-balanced Conformal Prediction \cite{angelopoulos2021gentle}.} In some applications, it is desirable to guarantee reliable uncertainty quantification not only marginally over the population but also within specific subpopulations.
This motivates group-balanced Conformal Prediction, where coverage is enforced separately for each group.
Formally, suppose the first component of the input $X_i$ indicates group membership, taking values in $\{1,\ldots,G\}$.
The goal is to construct a prediction set $\mathcal{C}$ such that, for every group $g\in\{1,\ldots,G\}$,
\begin{align}
\mathbb{P}\!\left( Y_{d+1}\in \mathcal{C}(X_{d+1}) \mid X_{d+1}\in g \right)\ge 1-\alpha.
\end{align}
Standard Conformal Prediction does not, in general, guarantee simultaneous coverage within each group.
To address this, group-balanced Conformal Prediction stratifies the calibration scores by group and computes group-specific conformal quantiles.
At test time, the prediction set is formed using the quantile corresponding to the group of $X_{d+1}$, yielding valid group-wise coverage under exchangeability.

\textit{Class-conditional Conformal Prediction \cite{angelopoulos2021gentle}.} In classification problems, it is often desirable to guarantee coverage separately for each ground-truth class rather than only marginally over the entire population.
This leads to class-conditional Conformal Prediction, which aims to ensure that prediction sets achieve the target coverage level within every class.
Formally, letting $\mathcal{Y}=\{1,\ldots, K\}$ denote the set of class labels, the goal is to construct a prediction set $\mathcal{C}$ such that
\begin{align}
\mathbb{P}\!\left( Y_{d+1}\in \mathcal{C}(X_{d+1}) \mid Y_{d+1}=k \right)\ge 1-\alpha,
\qquad \forall k\in\mathcal{Y}.
\end{align}
Unlike group-balanced Conformal Prediction, class membership is not directly observed at test time.
To address this, class-conditional Conformal Prediction stratifies the calibration scores by class and computes class-specific quantiles.
At inference time, candidate labels are included in the prediction set if their corresponding nonconformity scores fall below the class-specific threshold, yielding valid class-wise coverage under exchangeability.

\textit{Conformal Risk Control \cite{angelopoulos2022conformal}. }
Beyond bounding miscoverage, Conformal Prediction has been extended to control more general notions of error through conformal risk control.
Instead of enforcing a guarantee on the event $\{Y_{d+1}\notin\mathcal{C}(X_{d+1})\}$, this framework aims to bound the expected value of a user-specified loss function $\ell$, yielding guarantees of the form
\begin{align}
\mathbb{E}\!\left[\ell\!\left(\mathcal{C}(X_{d+1}),Y_{d+1}\right)\right]\le \alpha.
\end{align}
This formulation subsumes standard miscoverage as a special case and enables control over richer error metrics, such as the false negative rate or other task-dependent risks.
Conformal risk control operates by constructing a family of prediction sets indexed by a tuning parameter and selecting this parameter using calibration data so that the empirical risk satisfies a suitably adjusted threshold.
Under exchangeability and mild monotonicity assumptions on the loss, the resulting prediction set enjoys finite-sample, distribution-free guarantees on the target risk, making this approach particularly useful in applications where miscoverage alone does not capture the relevant notion of predictive error.

\textit{Conformal Prediction under Covariate Shift \cite{tibshirani2019conformal}. }
Classical Conformal Prediction relies on the assumption that calibration and test data are drawn from the same distribution.
In practice, this assumption is often violated due to covariate shift, where the marginal distribution of the inputs changes from $P$ to $P_{\mathrm{test}}$, while the conditional distribution $Y\mid X$ remains unchanged.
To handle this setting, weighted Conformal Prediction modifies the calibration procedure by reweighting calibration points according to their likelihood under the test distribution.
Specifically, calibration scores are weighted using importance weights
\begin{align}
w(x)=\frac{dP_{\mathrm{test}}(x)}{dP(x)},
\end{align}
so that points more representative of the test distribution receive a stronger influence.
The conformal threshold is then computed as a weighted quantile of the calibration scores.
Under correct specification of the weights and exchangeability after reweighting, this approach yields valid prediction sets for test points drawn from $P_{\mathrm{test}}$, extending Conformal Prediction to settings with covariate shift.

\textit{Conformal Prediction under Distribution Drift \cite{hu2025distribution}.}
Beyond covariate shift, Conformal Prediction has also been extended to settings with distribution drift, where the data distribution evolves gradually over time in a manner that may be unknown or difficult to estimate.
In this setting, calibration data and test points are no longer identically distributed, but recent observations are typically more informative about the test distribution.
Weighted Conformal Prediction addresses this by assigning larger weights to more recent calibration scores and computing a weighted conformal quantile.
The resulting prediction set is constructed as usual by thresholding the nonconformity score.
Under mild assumptions, including bounded drift measured via total variation distance, this approach preserves approximate marginal coverage, with explicit bounds that depend on both the magnitude of the drift and the chosen weighting scheme.
This framework provides a principled way to maintain valid uncertainty quantification in non-stationary environments such as time-series and streaming data.

\subsection{Evaluation}\label{app: ConfMeasure}
\textbf{Coverage. }Coverage is the central criterion for evaluating the quality of a Conformal Prediction method, as it quantifies the reliability of the resulting prediction sets. Formally, for a target miscoverage level $\gamma \in (0,1)$, Conformal Prediction aims to ensure that the prediction set $\mathcal{C}(X)$ satisfies
\begin{align}
\mathbb{P}\!\left( Y_{\mathrm{test}} \in \mathcal{C}(X_{\mathrm{test}}) \right) \ge 1 - \gamma,
\end{align}
independently of the underlying data distribution. In practice, empirical coverage on a test set $\{(X_i^{\mathrm{test}}, Y_i^{\mathrm{test}})\}_{i=1}^{n_{\mathrm{test}}}$ is estimated as
\begin{align}
\widehat{\mathrm{Cov}}
= \frac{1}{n_{\mathrm{test}}}
\sum_{i=1}^{n_{\mathrm{test}}}
\mathbf{1}\!\left\{ Y_i^{\mathrm{test}} \in \mathcal{C}(X_i^{\mathrm{test}}) \right\},
\end{align}
which measures the frequency with which the true outcomes are contained in the predicted sets. Coverage is essential because it provides a rigorous, interpretable guarantee on uncertainty quantification: a method achieving coverage close to the nominal level $1-\gamma$ can be trusted to capture the true outcome with the prescribed probability, whereas under-coverage indicates overconfident and potentially unsafe predictions.

\textbf{Prediction Set Size \cite{dhillon2024expected}. }While coverage ensures reliability, the size of the prediction set determines the practical usefulness and interpretability of Conformal Prediction. Among methods achieving the same coverage level, smaller prediction sets are preferred, as they provide more informative and less conservative uncertainty quantification. The notion of size depends on the output space: for classification, it is typically measured by the cardinality of the label set, while for regression, it may correspond to interval length or volume. Formally, the empirical average size of the prediction sets on a test set $\{X_i^{\mathrm{test}}\}_{i=1}^{n_{\mathrm{test}}}$ is given by
\begin{align}
\widehat{\mathrm{Size}}
= \frac{1}{n_{\mathrm{test}}}
\sum_{i=1}^{n_{\mathrm{test}}}
\lvert \mathcal{C}(X_i^{\mathrm{test}}) \rvert,
\end{align}
where $\lvert \mathcal{C}(X) \rvert$ denotes the cardinality or volume of the prediction set. Evaluating prediction set size alongside coverage highlights the fundamental coverage-size trade-off ina  Conformal Prediction: overly large sets may trivially achieve coverage but lack informativeness, whereas excessively small sets risk violating the coverage guarantee.

\textbf{Adaptability.}
Beyond marginal coverage and average prediction set size, a desirable Conformal Prediction procedure should be adaptive, producing small prediction sets for easy inputs and larger sets for harder or more uncertain inputs in a manner that faithfully reflects the underlying model uncertainty. Adaptability captures whether uncertainty is allocated heterogeneously across the input space rather than uniformly, avoiding overly conservative behavior on simple instances. A common way to evaluate adaptability is to partition the test data into bins based on a proxy for difficulty or uncertainty, such as ranges of the test input, ranges of the predicted output, model confidence scores, or the resulting prediction set size, and then compute empirical coverage separately within each bin. Formally, for a partition $\{\mathcal{B}_k\}_{k=1}^K$ of the test set, one evaluates
\begin{align}
\widehat{\mathrm{Cov}}_k
= \frac{1}{|\mathcal{B}_k|}
\sum_{i \in \mathcal{B}_k}
\mathbf{1}\!\left\{ Y_i^{\mathrm{test}} \in \mathcal{C}(X_i^{\mathrm{test}}) \right\},
\end{align}
and assesses whether coverage remains close to the nominal level across all bins. Consistent coverage across bins indicates that the method adapts appropriately to varying levels of uncertainty, while systematic deviations reveal a lack of adaptivity or localized miscalibration.
\begin{remark}[Evaluation Procedure for task \textbf{(III)}]
    In our setting, both inputs and outputs are probability distributions, which makes it difficult to define meaningful bins directly in the input or output space. As a result, we evaluate adaptability by partitioning the test samples according to the size of their conformal prediction sets, which serves as an intrinsic, model-agnostic proxy for predictive uncertainty. Specifically, we sort the test instances by prediction set size and divide them into bins such that each bin contains approximately the same number of test samples. For each bin, we compute the empirical coverage separately and examine whether it remains close to the nominal level. This evaluation allows us to assess whether larger prediction sets are indeed allocated to harder instances while maintaining valid coverage across different uncertainty regimes. We visualize this behavior by plotting the empirical coverage across a selected number of bins. Our approach follows that of \cite{thurin2025optimal}.
\end{remark}
\subsection{General Applications}
Conformal Prediction has been successfully applied across a wide range of learning problems by augmenting point predictions with statistically valid prediction sets. Below, we briefly describe several representative applications, highlighting how Conformal Prediction is used, the form of the prediction sets, and the type of uncertainty control involved.

\textbf{Multilabel Classification \cite{papadopoulos2014cross}.}
In multilabel classification, Conformal Prediction is used to output a set of labels for each input rather than a single label vector. Prediction sets are constructed by conformalizing label-wise or joint nonconformity scores derived from probabilistic classifiers, ensuring that the true label set (or each true label) is contained with high probability. These methods often emphasize risk control, such as controlling the false discovery rate or expected set size, and can be adapted to label-conditional settings.

\textbf{Disease (Sickness) Classification \cite{vazquez2022conformal}.}
For disease classification tasks, Conformal Prediction produces a set of plausible diagnoses for a patient. The prediction set consists of disease labels whose scores satisfy a calibrated conformity threshold, guaranteeing that the true diagnosis is included with a prescribed probability. This setting primarily focuses on risk-sensitive decision making, allowing clinicians to identify ambiguous cases and defer decisions when uncertainty is high.

\textbf{Polyp Segmentation \cite{mossina2025conformal}.}
In polyp segmentation, Conformal Prediction is applied to construct uncertainty-aware segmentation regions rather than a single binary mask. Prediction sets may take the form of pixel-wise confidence regions or spatial bands around predicted boundaries, providing coverage guarantees for the true polyp region. These sets are particularly useful for safety-critical applications, where missing polyp tissue carries a significant risk.

\textbf{Weather Prediction \cite{mortier2025evaluating}.}
In weather forecasting, Conformal Prediction is used to generate prediction intervals or sets for continuous variables such as temperature or precipitation. These sets are typically constructed by conformalizing regression residuals or ensemble forecasts and provide valid coverage without strong distributional assumptions. This application often addresses distribution shift across time or locations through split or adaptive calibration.

\textbf{Quantile Regression \cite{romano2019conformalized}.}
In quantile regression, Conformal Prediction is used to transform estimated conditional quantiles into valid prediction intervals. By conformalizing lower and upper quantile estimates, the resulting intervals achieve guaranteed marginal coverage even under model misspecification. This approach is particularly effective in heteroskedastic settings and focuses on uncertainty calibration rather than pointwise accuracy.

\textbf{Toxic Online Comment Identification \cite{villate2024collaborative}.}
For toxic content detection, Conformal Prediction constructs prediction sets over toxicity labels to identify uncertain or borderline cases. When the conformal set contains multiple labels, the system may defer judgment or flag content for human review. This application emphasizes selective risk control, reducing harmful misclassifications in automated moderation systems.

\textbf{Selective Classification \cite{boursinos2022selective}.}
Selective classification naturally aligns with Conformal Prediction by allowing models to abstain when prediction sets are non-singleton or empty. Predictions are made only when the conformal set contains a single label, while abstention occurs otherwise. This framework provides explicit guarantees on the error rate of accepted predictions and enables principled risk-coverage trade-offs.

\subsection{Applications in DDR}\label{app: apliinDDR}
While conformal prediction (CP) has been applied broadly across many real-world settings, this work focuses specifically on CP for distribution-to-distribution regression (DDR). For completeness, this section briefly discusses several representative applications of CP within the DDR framework.

\textbf{Model Reliability Quantification.}
A practitioner can use the size of the prediction set to decide whether to trust the model, defer to a human expert, or reject the input altogether. This type of decision rule is particularly important in high-stakes domains such as healthcare, genomics, and scientific discovery, where acting on an uncertain prediction can lead to significant consequences. For instance, in medical or gene expression analysis~\cite{oliva2013distribution}, a small prediction set over distributions indicates that the model has identified a well-concentrated and reliable range of plausible biological behaviors, allowing automated conclusions or downstream analysis. In contrast, a large or diffuse prediction set signals substantial uncertainty or heterogeneity in the data, prompting further experimental validation or expert review. In this sense, prediction sets for DDR provide a principled way to assess the reliability of distributional predictions and guide downstream decisions accordingly.

\textbf{Robust Decision Making.}
Instead of relying on a single estimated distribution, the practitioner can make decisions that remain reliable across all plausible distributions in the prediction set. Concretely, given $\mathcal C(X)$, one can evaluate a decision, such as pricing, allocation, or investment choice, under each candidate distribution and select an option that performs well in the worst-case or conservative scenario. For example, in house price prediction~\cite{chen2023wasserstein}, rather than committing to a single estimated price distribution, a buyer or investor can consider all plausible distributions in $\mathcal C(X)$ and choose a strategy that avoids overpaying even in the most adverse case. This is particularly important because, in the DDR setting where outputs are complex distributions, model uncertainty and misspecification are more likely; prediction sets therefore provide a principled way to guard against such risks and support more robust decision-making.

\textbf{Hypothesis Testing.}
In many applications, the goal is not merely to produce a prediction, but to understand structural properties of the output distribution, such as concentration, multimodality, or tail behavior. However, when data are limited, validating such properties from a single-point estimate of the distribution is inherently unreliable, as small-sample noise or model misspecification can significantly distort these characteristics. In DDR, this challenge is even more pronounced because of the complexity of the distributional outputs. A prediction set $\mathcal C(X)$ addresses this by providing a collection of plausible distributions with coverage guarantees, allowing practitioners to assess whether a property holds consistently across the set rather than relying on a single estimate. For example, in age prediction~\cite{chen2023wasserstein}, one may wish to determine whether the predicted age distribution is concentrated around a single age group or exhibits multiple plausible age ranges. If this property holds across all distributions in $\mathcal C(X)$ over many $X$, the conclusion is robust; otherwise, it indicates that the data is insufficient to support a definitive claim. In this sense, prediction sets for DDR act as a form of uncertainty-aware data augmentation, enabling more reliable validation of output properties in the complex and low-data setting of DDR.

In summary, conformal prediction in the standard setting has been widely adopted not merely as a predictive tool but as a framework that enables a range of downstream statistical tasks, including reliability-aware decision making, robust optimization, and hypothesis validation. In the DDR setting, these challenges become significantly more pronounced due to the increased complexity of distributional outputs and the higher risk of model misspecification. Our method extends conformal prediction to this setting, providing valid prediction sets for distributions and thereby ensuring that these downstream tasks and statistical objectives can still be carried out in a principled and reliable manner.

\section{Theoretical Proofs}
\subsection{Proof of the Marginal Coverage Guarantee - Theorem~\ref{thm: MarginCov}} \label{proof: MarginCov}

\begin{proof}
We have that $\{(X_i, Y_i)\}_{i=1}^{d_0}$ and $(X_0, Y_0)$ are exchangeable. Since $\tilde y_i$ (and hence $\theta(\tilde y_i)$) is a measurable function of the observed sample from $Y_i$, applying the same measurable map to each pair yields exchangeable scores $(s_0,s_1,\dots,s_{d_0})$.
Let $R$ denote the rank of $s_0$ among $(s_0,s_1,\dots,s_{d_0})$ after random tie-breaking. By exchangeability, $R$ is uniform on $\{1,\dots,d_0+1\}$. With $k=\lceil(d_0+1)(1-\gamma)\rceil$ and $\hat q=s_{(k)}$ computed from $\{s_i\}_{i=1}^{d_0}$, the conformal event is
\begin{align}
\theta(\tilde y_0)\in\mathcal E(X_0)\quad\text{if and only if}\quad s_0\le \hat q.
\end{align}
Under random tie-breaking, $\{s_0\le \hat q\}$ contains the event $\{R\le k\}$, hence
\begin{align}
\mathbb P\big(s_0\le \hat q\big)\ \ge\ \mathbb P(R\le k)\ =\ \frac{k}{d_0+1}\ \ge\ 1-\gamma.
\end{align}
If ties occur at the threshold, $\{s_0\le \hat q\}$ may include up to a tie block of size at most $\bar{d}$ beyond rank $k$. Since $R$ is uniform under random tie-breaking, this adds at most $\bar{d}/(d_0+1)$ probability mass, yielding
\begin{align}
\mathbb P\big(s_0\le \hat q\big)\ \le\ \frac{k}{d_0+1}+\dfrac{\bar{d}}{d_0+1}\ \le\ 1-\gamma+\dfrac{\bar{d}}{d_0+1}.
\end{align}
Therefore,
\begin{align}
1-\gamma \ \le\ \mathbb P\!\big(\theta(\tilde y_0)\in\mathcal E(X_0)\big)\ \le\ 1-\gamma+\dfrac{\bar{d}}{d_0+1}.
\end{align}
By construction of $\mathcal Q(X_0)$ via the truncated basis map $a\mapsto y_a$, we have
\begin{align}
\tilde y_0\in\mathcal Q(X_0)\quad\text{if and only if}\quad \theta(\tilde y_0)\in\mathcal E(X_0),
\end{align}
and thus the same bounds apply to $\mathbb P(\tilde y_0\in\mathcal Q(X_0))$.
\end{proof}
\subsection{Proof of the Prediction Set Volume - Theorem~\ref{thm: volMarg}}\label{proof of the Prediction Set Volume}
\begin{proof}
Let $t=|T_R|$ and enumerate $T_R=\{\alpha_1,\ldots,\alpha_t\}$. Define the finite-dimensional coefficient-to-function map
$
\Phi:\mathbb{R}^t\to \operatorname{span}\{\beta_\alpha:\alpha\in T_R\}
$
by
\begin{align}
\Phi(a)(z)
=
\sum_{\alpha\in T_R}a_\alpha\beta_\alpha(z).
\end{align}
By construction,
\begin{align}
\mathcal{Q}(X_0)=\Phi\big(\mathcal{E}(X_0)\big).
\end{align}
Since $\{\beta_\alpha\}_{\alpha\in T_R}$ is an orthonormal family in $L_2(\Lambda^l)$, the map $\Phi$ is a linear isometry from $\mathbb{R}^t$ equipped with the Euclidean norm to the finite-dimensional subspace $\operatorname{span}\{\beta_\alpha:\alpha\in T_R\}\subset L_2(\Lambda^l)$. Indeed, for any $a,b\in\mathbb{R}^t$,
\begin{align}
\|\Phi(a)-\Phi(b)\|_{L_2(\Lambda^l)}^2
&=
\left\|
\sum_{\alpha\in T_R}(a_\alpha-b_\alpha)\beta_\alpha
\right\|_{L_2(\Lambda^l)}^2  \\
&=
\sum_{\alpha\in T_R}(a_\alpha-b_\alpha)^2 \\
&=
\|a-b\|_2^2.
\end{align}
Therefore, the $t$-dimensional volume of $\mathcal{Q}(X_0)$ in the induced coefficient subspace is equal to the Lebesgue volume of $\mathcal{E}(X_0)$ in $\mathbb{R}^t$:
\begin{align}
\operatorname{Vol}\big(\mathcal{Q}(X_0)\big)
=
\operatorname{Vol}\big(\mathcal{E}(X_0)\big).
\end{align}

It remains to compute the volume of $\mathcal{E}(X_0)$. Let
$
m=\theta(\hat f(X_0)).
$
Since $g=\|\cdot\|_2$, the coefficient-space prediction set is
\begin{align}
\mathcal{E}(X_0)
=
\left\{
a\in\mathbb{R}^t:
\|W^{1/2}(m-a)\|_2\le \hat q
\right\}.
\end{align}
Thus $\mathcal{E}(X_0)$ is an ellipsoid centered at $m$. Under the change of variables
\begin{align}
u=W^{1/2}(a-m),
\qquad
a=m+W^{-1/2}u,
\qquad
da=\frac{1}{\sqrt{\det W}}\,du.
\end{align}
The defining constraint becomes
\begin{align}
\|u\|_2\le \hat q.
\end{align}
Hence,
\begin{align}
\operatorname{Vol}\big(\mathcal{E}(X_0)\big)
&=
\frac{1}{\sqrt{\det W}}
\operatorname{Vol}\left(\{u\in\mathbb{R}^t:\|u\|_2\le \hat q\}\right) \\
&=
\frac{1}{\sqrt{\det W}}
\cdot
\frac{\pi^{t/2}\hat q^t}{\Gamma\left(\frac{t}{2}+1\right)}.
\end{align}
Combining this with
$
\operatorname{Vol}\big(\mathcal{Q}(X_0)\big)
=
\operatorname{Vol}\big(\mathcal{E}(X_0)\big)
$,
we obtain
\begin{align}
\operatorname{Vol}\big(\mathcal{Q}(X_0)\big)
=
\frac{\pi^{t/2}\hat q^t}{\Gamma\left(\frac{t}{2}+1\right)\sqrt{\det W}}
=
\frac{(\sqrt{\pi}\hat q)^t}{\Gamma\left(\frac{t}{2}+1\right)\sqrt{\det W}}.
\end{align}
This completes the proof.
\end{proof}
\subsection{Proof of the Marginal Coverage Law Given the Calibration Set -  Theorem~\ref{thm: marg gamma}}
 \label{proof: T4.4}
\begin{proof}
For each calibration pair $(X_i^c, Y_i^c)$, $i = 1, \ldots, d_0$, let
\begin{align}
S_i = s(\hat f(X_i^c), Y_i^c)
\end{align}
denote its conformal score, and let
$
S_0 = s(\hat f(X_0), Y_0)
$
be the score of the test pair $(X_0, Y_0)$. By the exchangeability of the data,
the random variables $S_0, S_1, \ldots, S_{d_0}$ are i.i.d.\ from some
distribution with a continuous cumulative distribution function $F$. Let
\begin{align}
S_{(1)} \le S_{(2)} \le \cdots \le S_{(d_0)}
\end{align}
denote the order statistics of $S_1, \ldots, S_{d_0}$, and define the
threshold
$
T := S_{(d_0 + 1 - \ell)}.
$
By the definition of the Conformal Prediction set $\mathcal{Q}(X_0)$, we have
\begin{align}
\tilde y_0 \in \mathcal{Q}(X_0)
\quad \text{if and only if} \quad
S_0 \le T.
\end{align}

Hence, conditional on the calibration sample
$\{(X_i^c, Y_i^c)\}_{i=1}^{d_0}$,
\begin{align}
\mathbb{P}\!\left(
\tilde y_0 \in \mathcal{Q}(X_0)
\;\middle|\;
\{(X_i^c, Y_i^c)\}_{i=1}^{d_0}
\right)
&=
\mathbb{P}\!\left(
S_0 \le T \mid \{S_i\}_{i=1}^{d_0}
\right)
\\&= F(T).
\end{align}
Thus, the conditional coverage probability is the random variable $F(T)$. Next, apply the probability integral transform. Define
$
U_i := F(S_i), \qquad i = 1, \ldots, d_0.
$
Since $F$ is continuous and $S_i \sim F$, it follows that
$U_i \sim \mathrm{Unif}(0,1)$ i.i.d. Let
\begin{align}
U_{(1)} \le \cdots \le U_{(d_0)}
\end{align}
denote the order statistics of $\{U_i\}_{i=1}^{d_0}$. Because $F$ is
monotone increasing and $T = S_{(d_0 + 1 - \ell)}$, we have
\begin{align}
F(T)
= F\bigl(S_{(d_0 + 1 - \ell)}\bigr)
= U_{(d_0 + 1 - \ell)}.
\end{align}

A standard result for order statistics of i.i.d.\ $\mathrm{Unif}(0,1)$
Random variables states that
\begin{align}
U_{(k)} \sim \mathrm{Beta}(k,\, d_0 + 1 - k),
\qquad k = 1, \ldots, d_0.
\end{align}
Taking $k = d_0 + 1 - \ell$ yields
\begin{align}
F(T) = U_{(d_0 + 1 - \ell)}
\sim \mathrm{Beta}(d_0 + 1 - \ell,\, \ell).
\end{align}

Since $F(T)$ is precisely the conditional coverage probability, we conclude
that
\begin{align}
\mathbb{P}\!\left(
\tilde y_0 \in \mathcal{Q}(X_0)
\;\middle|\;
\{(X_i^c, Y_i^c)\}_{i=1}^{d_0}
\right)
\sim \mathrm{Beta}(d_0 + 1 - \ell,\, \ell),
\end{align}
which completes the proof.
\end{proof}

\subsection{Proof of Distribution of Coverage Given the Calibration Set - Corollary \ref{coro: binombetamargin}}\label{proof: binombetamargin}
\begin{proof}
Let $\ell := \lfloor (d_0+1)\gamma \rfloor$, and let
$\{(X_0^{(j)},Y_0^{(j)})\}_{j=1}^t$ be an exchangeable collection of test pairs, independent of the calibration sample.
For each calibration pair $(X_i^c,Y_i^c)$, $i=1,\ldots,d_0$, define the score
$
S_i := s(X_i^c,Y_i^c),
$
and for each test pair define
\begin{align}
S_0^{(j)} := s\bigl(X_0^{(j)},Y_0^{(j)}\bigr),\qquad j=1,\ldots,t.
\end{align}
Let $S_{(1)}\le \cdots \le S_{(d_0)}$ be the order statistics of $S_1,\ldots,S_{d_0}$ and set
$
T := S_{(d_0+1-\ell)}.
$
By definition of $\mathcal Q(\cdot)$, for each $j$,
\begin{align}
\tilde y_0^{(j)} \in \mathcal Q(X_0^{(j)}) \quad \text{if and only if} \quad S_0^{(j)} \le T,
\end{align}
and therefore
$
I = \sum_{j=1}^t \mathbf 1\!\left\{S_0^{(j)} \le T\right\}.
$
Conditioning on the calibration data $\{(X_i^c,Y_i^c)\}_{i=1}^{d_0}$, so that $T$ is fixed.
By the same exchangeability/continuity argument as in the proof of \cref{thm: marg gamma}, the calibration scores and each test score are i.i.d.\ from some continuous CDF $F$, hence for every $j$,
\begin{align}
\mathbb P\!\left(S_0^{(j)} \le T \,\middle|\,\{(X_i^c,Y_i^c)\}_{i=1}^{d_0}\right) = F(T).
\end{align}
Moreover, since the test pairs are exchangeable\ and independent of the calibration sample, the indicators
$\mathbf 1\{S_0^{(j)}\le T\}$ are conditionally exchangeable\ given $\{S_i\}_{i=1}^{d_0}$, and thus
\begin{align}
I \,\big|\,  \{(X_i^c,Y_i^c)\}_{i=1}^{d_0},\,F(T)
\;\sim\;
\mathrm{Binomial}\bigl(t,\,F(T)\bigr).
\end{align}

Finally, by \cref{thm: marg gamma} we have
\begin{align}
F(T)
=
\mathbb P\!\left(\tilde y_0 \in \mathcal Q(X_0)\,\middle|\, \{(X_i^c,Y_i^c)\}_{i=1}^{d_0}\right)
\sim
\mathrm{Beta}(d_0+1-\ell,\ell).
\end{align}
Integrating the conditional binomial law over the random success probability $F(T)$ yields the Beta-Binomial mixture,
\begin{align}
I \,\big|\, \{(X_i^c,Y_i^c)\}_{i=1}^{d_0}
\;\sim\;
\mathrm{BetaBinomial}\!\left(t,\, d_0+1-\ell,\, \ell\right),
\end{align}
which completes the proof.
\end{proof}

\subsection{Proof of Asymptotic Conditional Coverage - Theorem~\ref{thm: CondCov}} \label{proof: CondCov}
\begin{proof}
Let $a_0$ be the coefficient vector associated with $\tilde y_0$, and define
\begin{align}
S_0 := g\!\left(W^{1/2}\bigl(\theta(\hat f(X_0)) - a_0\bigr)\right).
\end{align}
By the definition of $\mathcal Q(X_0)$, the event $\{\tilde y_0 \in \mathcal Q(X_0)\}$
is equivalent to $\{S_0 \le \hat q_{X_0}\}$. Hence, the desired conclusion
can be rewritten as
\begin{align}
\mathbb P\bigl(S_0 \le \hat q_{X_0} \mid X_0\bigr)
\xrightarrow[]{\mathbb{P}} 1 - \gamma
\qquad \text{as } d_0 \to \infty.
\label{eq:target-S0}
\end{align}

To establish \eqref{eq:target-S0}, we use the regularity assumptions on the
conditional distribution of the score $S$ given $X$, stated above.
For $x$ in the support of $X$, denote by
\begin{align}
F_x(s) := \mathbb P(S \le s \mid X = x)
\end{align}
the conditional cumulative distribution function of $S$ given $X = x$, and
write $q_{1-\gamma}(x)$ for the $(1-\gamma)$-quantile of this distribution,
\begin{align}
q_{1-\gamma}(x)
:= \inf\{s \in \mathbb R : F_x(s) \ge 1 - \gamma\}.
\end{align}
By the absolute continuity and the local lower and upper bounds on the
conditional density $p_S(\cdot \mid x)$ (The above assumption), $F_x$ is strictly
increasing and continuous in a neighborhood of $q_{1-\gamma}(x)$, and thus
$F_x(q_{1-\gamma}(x)) = 1 - \gamma$.

Let $\hat F_k(\cdot \mid X_0)$ denote the $k$-nearest-neighbor empirical
conditional CDF built from the scores
$\{ s_i^{X_0} : i \in \mathcal I_{X_0} \}$,
\begin{align}
\hat F_k(s \mid X_0)
:= \frac{1}{k} \sum_{i \in \mathcal I_{X_0}}
\mathbf 1\{ s_i^{X_0} \le s \}, \qquad s \in \mathbb R.
\end{align}
By construction, letting
\begin{align}
r_k := \left\lceil (k+1)(1-\gamma)\right\rceil,
\end{align}
the local conformal threshold $\hat q_{X_0}$ is the $r_k$-th order statistic
of the local scores. Since
\begin{align}
\frac{r_k}{k} \longrightarrow 1-\gamma
\qquad\text{as } k\to\infty,
\end{align}
the consistency condition in Theorem~\ref{thm: CondCov}, together with the
continuity and strict monotonicity of $F_{X_0}$ in a neighborhood of
$q_{1-\gamma}(X_0)$, yields
\begin{align}
\hat q_{X_0}
\xrightarrow{\mathbb P}
q_{1-\gamma}(X_0).
\label{eq:qhat-consistency}
\end{align}
In particular, by the continuity and strict monotonicity of $F_{X_0}$ in a
neighborhood of $q_{1-\gamma}(X_0)$, this pointwise convergence implies
\begin{align}
\hat q_{X_0}
\xrightarrow{\mathbb P} q_{1-\gamma}(X_0)
\qquad \text{as } d_0 \to \infty.
\label{eq}
\end{align}

Finally, observe that
\begin{align}
\mathbb P(S_0 \le \hat q_{X_0} \mid X_0)
= F_{X_0}(\hat q_{X_0}).
\end{align}
By \eqref{eq:qhat-consistency} and the continuity of $F_{X_0}$ at
$q_{1-\gamma}(X_0)$, we obtain
\begin{align}
F_{X_0}(\hat q_{X_0})
\xrightarrow{\mathbb P}
F_{X_0}(q_{1-\gamma}(X_0))
= 1 - \gamma.
\end{align}
This establishes \eqref{eq:target-S0}, and therefore
\begin{align}
\mathbb P\bigl(\tilde y_0 \in \mathcal Q(X_0) \mid X_0\bigr)
= \mathbb P\bigl(S_0 \le \hat q_{X_0} \mid X_0\bigr)
\xrightarrow{\mathbb P} 1 - \gamma,
\end{align}
which completes the proof.
\end{proof}

\subsection{Proof of the Conditional Coverage Law under Local Calibration - Theorem~\ref{thm: local_gamma}}\label{proof: localgamma}
\begin{proof}
Let $S_{(1)} \le \cdots \le S_{(k)}$ denote the order statistics of
$\{ S_i : i \in \mathcal I_{X_0} \}$.
Fix $X_0$ and the calibration sample.
Conditionally on $\{(X_i^c, Y_i^c)\}_{i=1}^{d_0}$ and $X_0$,
the neighbor index set $\mathcal I_{X_0}$ is fixed, and hence the conformal
threshold
\begin{align}
T := S_{(k+1-\ell)}
\end{align}
is deterministic.
By the definition of $\mathcal Q(X_0)$,
\begin{align}
\tilde y_0 \in \mathcal Q(X_0)
\quad \text{if and only if} \quad
S_0 \le T.
\end{align}
Therefore,
$
\mathbb P\!\left(
\tilde y_0 \in \mathcal Q(X_0)
\ \middle|\
\{(X_i^c, Y_i^c)\}_{i=1}^{d_0},\ X_0
\right)
$ is equal to $
\mathbb P\!\left(
S_0 \le T
\ \middle|\
\{ S_i \}_{i \in \mathcal I(X_0)},\ X_0
\right).
$

Under Assumption \ref{ass: A3}, the conditional distribution
$S_0 \mid (X_0, \mathcal I_{X_0})$ follows $F_{X_0}$ and is independent of
$\{ S_i \}_{i \in \mathcal I_{X_0}}$.
Hence, the right-hand side equals
$
F_{X_0}(T).
$

Applying the probability integral transform, define
\begin{align}
U_i := F_{X_0}(S_i), \qquad i \in \mathcal I_{X_0}.
\end{align}
Since $F_{X_0}$ is continuous, the random variables
$\{ U_i : i \in \mathcal I_{X_0} \}$ are i.i.d.\ $\mathrm{Unif}(0,1)$.
Moreover, their order statistics satisfy
\begin{align}
F_{X_0}(T)
=
F_{X_0}\!\bigl(S_{(k+1-\ell)}\bigr)
=
U_{(k+1-\ell)}.
\end{align}
A standard result on order statistics implies that
$U_{(j)} \sim \mathrm{Beta}(j,\, k+1-j)$ for $j = 1, \ldots, k$.
Taking $j = k+1-\ell$ yields
\begin{align}
F_{X_0}(T)
\sim \mathrm{Beta}(k+1-\ell,\ \ell),
\end{align}
which is exactly the claimed distribution for the conditional coverage
probability.
\end{proof}
\section{Experimental Details}\label{app: expdet}
\subsection{Simulation Study}\label{App: Símtudy}
\textbf{Experimental setup.}
Unless otherwise stated, for the Truncated Gaussian and Mixture of Betas
datasets, we generate $1000$ training, $1000$ calibration,
and $1000$ test distribution pairs.
Each input and output distribution is observed through
$1000$ i.i.d. samples, which are then used to construct empirical
distributions and their corresponding coefficient representations.

For a nonparametric distribution regression model that requires bandwidth selection, we further generate
$200$ distribution pairs that are used exclusively for cross-validation.
The bandwidth is selected by minimizing the empirical $\ell_2$ error between
predicted and observed output coefficients on this validation set.

All output distributions are represented using a
Fourier orthonormal basis.
For one-dimensional synthetic datasets, the basis is truncated at
$R = 10$.
For the Dark Matter datasets, the Fourier bases are truncated at
$R = 5$ for the 3D setting and $R = 3$ for the 6D setting.
The Sobolev weighting parameters are fixed to $\nu = 1$ and $\sigma = 1$
throughout.

We evaluate on three tasks. \\
\textbf{(I) Efficiency-coverage trade-off:}
the target coverage level $1-\gamma$ is varied from $0.70$ to $0.96$, and \textit{Efficiency} is computed by averaging over calibration threshold $\hat q$ across test inputs (for marginal CP, $\hat q$ is identical for all inputs). Both marginal and conditional Conformal Prediction are evaluated.
For conditional Conformal Prediction, the neighborhood size is fixed to
$K = 500$ nearest neighbors, and the prediction set volumes are averaged
over all test points.

\textbf{(II) Beta-Binomial verification:}
for the Truncated Gaussian and Mixture of Betas datasets, a fixed pool
of $5000$ distribution pairs is generated once.
For the Dark Matter datasets, a fixed pool of $2300$ distribution pairs
is used.
In each repetition, $1000$ samples are randomly selected for calibration
and $1000$ samples for testing.
Empirical coverage is computed on the test set, and this resampling procedure is
repeated $50{,}000$ times to obtain the empirical distribution of coverage
values.

\textbf{(III) Conditional coverage:}
test inputs are sorted by prediction set size and partitioned into groups of
equal cardinality from smallest to largest set size, and empirical coverage is
computed within each group.
Conditional Conformal Prediction again uses $K = 500$ nearest neighbors.

\textbf{Visualization:}
$100$ coefficient vectors are uniformly sampled from the $\ell_2$-ball
centered at the predicted output coefficients, with a radius determined by the
conformal threshold, and mapped back to the distribution space via the truncated
basis expansion.

The data generation procedures for each dataset are described below.

\subsubsection{Truncated Gaussians}
\textbf{Data generation.}
We consider one-dimensional probability distributions supported on $[0,1]$. 
For each dataset instance $i$, we generate an input-output distribution pair $(\mu_i, \nu_i)$ as follows.

\textbf{Input distribution.}
First, we draw the mixture parameters independently:
\begin{center}
$
\mu_{i,1}, \mu_{i,2} \sim \mathrm{Unif}[0,1],
\quad
\sigma_{i,1}, \sigma_{i,2} \sim \mathrm{Unif}[0.05, 0.1].
$
\end{center}

The input density $p_i$ is defined as an equally weighted mixture of two Truncated Gaussian densities:
\begin{center}
$
p_i(x)
=
\frac{1}{2} g(x;\mu_{i,1},\sigma_{i,1})
+
\frac{1}{2} g(x;\mu_{i,2},\sigma_{i,2}),
\quad x \in [0,1].
$
\end{center}

Here, $g(\cdot;\mu,\sigma)$ denotes the Gaussian density with mean $\mu$ and standard deviation $\sigma$, truncated to $[0,1]$, defined as
\begin{center}
$
g(x;\mu,\sigma)
=
\frac{1}{\sigma}
\frac{
\phi\!\left(\frac{x-\mu}{\sigma}\right)
}{
\Phi\!\left(\frac{1-\mu}{\sigma}\right)
-
\Phi\!\left(\frac{-\mu}{\sigma}\right)
},
\quad x \in [0,1],
$
\end{center}
where $\phi$ and $\Phi$ denote the standard normal probability density function and cumulative distribution function, respectively.

\textbf{Output distribution.}
The corresponding output density $q_i$ is defined by reflecting the mixture means around $1/2$, while keeping the same standard deviations:
\begin{center}
$
q_i(x)
=
\frac{1}{2} g(x;1-\mu_{i,1},\sigma_{i,1})
+
\frac{1}{2} g(x;1-\mu_{i,2},\sigma_{i,2}),
\quad x \in [0,1].
$
\end{center}

\textbf{Sampling.}
Each input and output distribution is observed only through i.i.d.\ samples:
\begin{center}
$
\{X_{ij}\}_{j=1}^{m_i} \stackrel{\mathrm{i.i.d.}}{\sim} p_i,
\qquad
\{Y_{ij}\}_{j=1}^{n_i} \stackrel{\mathrm{i.i.d.}}{\sim} q_i.
$
\end{center}
\begin{figure}[t]
    \centering
    \includegraphics[width=\linewidth]{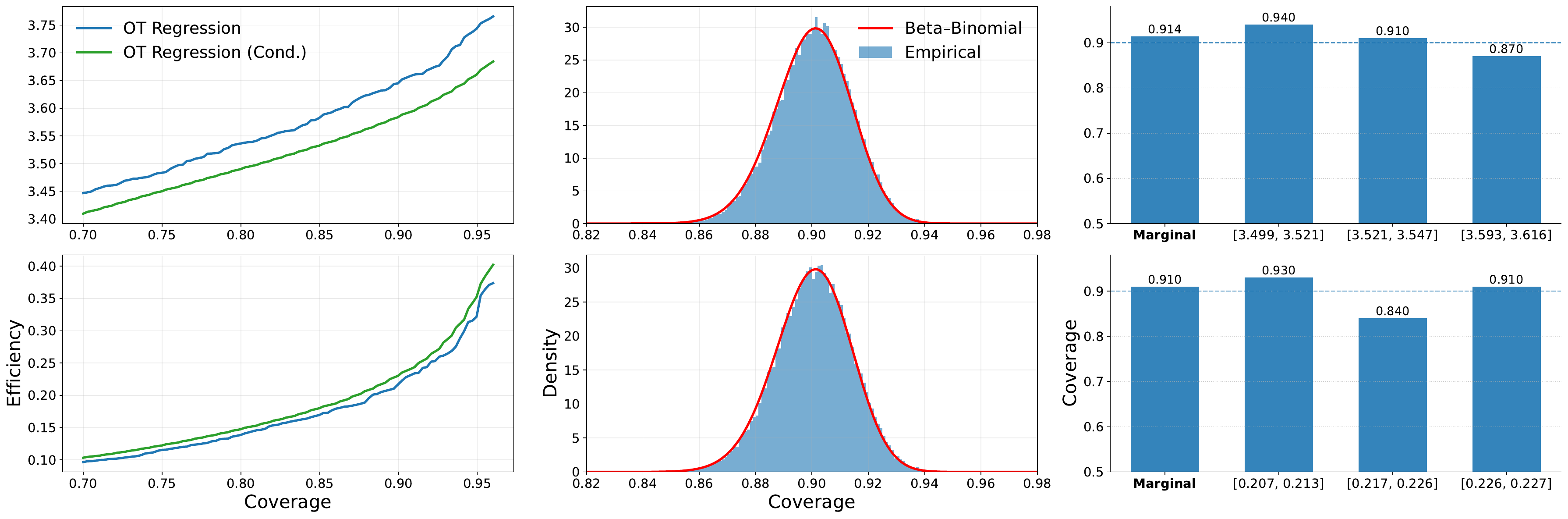}
    \caption{Experiments on Truncated Gaussian (first row) and Mixture of Betas (second row) for task \textbf{(I)} (first column), \textbf{(II)} (second column), and \textbf{(III)} (third column) using the OTR base model.}
    \label{fig:GaussBetaOTR}
\end{figure}

Sampling from the mixture is performed by first drawing a Bernoulli variable with probability $1/2$ to select a component, followed by sampling from the corresponding Truncated Gaussian distribution.

\subsubsection{Mixture of Betas}
\textbf{Data generation.}
We consider one-dimensional probability distributions supported on $[0,1]$ constructed as mixtures of Beta distributions combined with random transport-based noise. 
For each dataset instance $i$, we generate an input-output distribution pair $(\mu_i, \nu_i)$ as follows.

\textbf{Input distribution.}
First, we generate the input distribution $\mu_i$ as a mixture of three Beta components:
\begin{center}
$
\mu_i(dx)
=
\sum_{j=1}^{3} \pi_j \,
\mathrm{Beta}(x;\alpha_{i,j},\beta_{i,j}) \, dx,
\quad x \in [0,1],
$
\end{center}
where the mixture weights $\{\pi_j\}_{j=1}^3$ are fixed and satisfy $\sum_{j=1}^3 \pi_j = 1$. 
In all experiments, we set $\pi = (0.3, 0.4, 0.3)$.

The shape parameters of the Beta components are drawn independently as
\begin{center}
$
\alpha_{i,j} \sim \mathrm{Unif}[1,10],
\qquad
\beta_{i,j} \sim \mathrm{Unif}[1,10],
\quad j = 1,2,3.
$
\end{center}

Given a fixed sample size $m_i$, the input distribution is observed through i.i.d.\ samples
\begin{center}
$
\{X_{ij}\}_{j=1}^{m_i} \stackrel{\mathrm{i.i.d.}}{\sim} \mu_i.
$
\end{center}

\textbf{Regression operator.}
The regression operator is defined as a deterministic transport map followed by a random perturbation. 
Specifically, we define the base regression map $T_0 : [0,1] \to [0,1]$ as
\begin{center}
$
T_0(x) = \zeta_4(x),
\qquad
\zeta_k(x) = x - \frac{\sin(\pi k x)}{|k|\pi},
$
\end{center}
which is a strictly increasing and smooth map satisfying $T_0(0)=0$ and $T_0(1)=1$.

\textbf{Random error maps.}
To model stochastic variability, we introduce random transport maps $T_{\varepsilon, i}$. 
Let $J \ge 2$ and let
\begin{center}
$
0 = U_{(0)} < U_{(1)} < \cdots < U_{(J-1)} < U_{(J)} = 1
$
\end{center}
be the order statistics of $J-1$ i.i.d.\ random variables drawn from $\mathrm{Unif}[0,1]$.
For each interval $[U_{(j)}, U_{(j+1)}]$, we sample an integer-valued random variable $K_j$
from a symmetric distribution around zero. In our experiments, $K_j$ is supported
on a finite set of integers.

The random error map is then defined as a piecewise transport map:
\begin{center}
$
T_{\varepsilon,i}(x)
=
\sum_{j=0}^{J-1}
\mathbf{1}_{[U_{(j)},\,U_{(j+1)}]}(x)\,
\xi\!\left(U_{(j)}, U_{(j+1)}, K_j\right)(x),
$
\end{center}
where $\xi(\cdot)$ denotes the affine rescaling of $\zeta_{K_j}$ to the interval $[U_{(j)}, U_{(j+1)}]$. 
By symmetry of $K_j$, the random maps satisfy
\begin{center}
$
\mathbb{E}[T_{\varepsilon,i}(x)] = x,
\quad \forall x \in [0,1].
$
\end{center}
\begin{figure}[t]
    \centering
    \includegraphics[width=\linewidth]{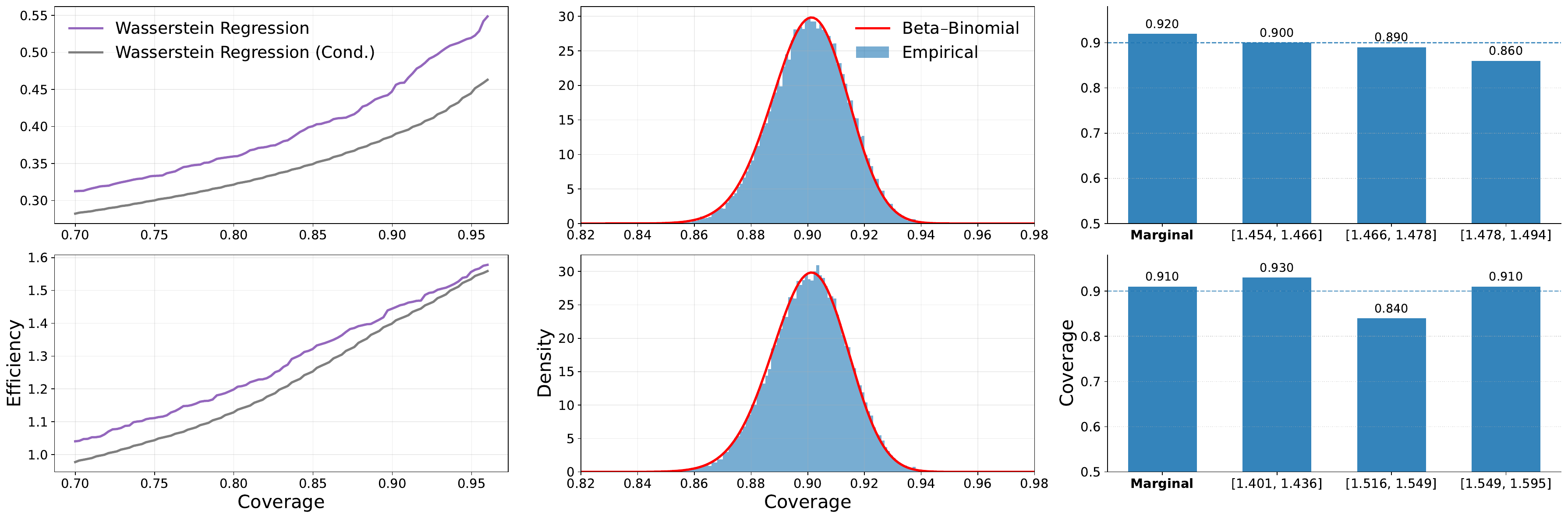}
    \caption{Experiments on Truncated Gaussian (first row) and Mixture of Betas (second row) for task \textbf{(I)} (first column), \textbf{(II)} (second column), and \textbf{(III)} (third column) using the WR base model.}
    \label{fig:GaussBetaWR}
\end{figure}
\textbf{Output distribution.}
The output distribution $\nu_i$ is defined as the push-forward of $\mu_i$ through the composed transport map:
\begin{center}
$
\nu_i = T_{\varepsilon,i\#}\, T_{0\#}\mu_i.
$
\end{center}

In practice, given samples $\{X_{ij}\}_{j=1}^{m_i}$, the response samples are generated as
\begin{center}
$
Y_{ij} = T_{\varepsilon,i}\!\left(T_0(X_{ij})\right),
\quad j=1,\dots,m_i,
$
\end{center}
followed by truncation to $[0,1]$ if necessary.

This construction yields nonlinear input-output relationships with distribution-dependent noise, providing a challenging benchmark for distribution-to-distribution regression and Conformal Prediction methods.

\subsubsection{Dark Matter Simulation Data}
\textbf{Data generation.}
We construct synthetic distribution-to-distribution regression datasets based on cosmological Dark Matter simulations generated with the \textsc{FastPM} framework \cite{feng2016fastpm}, following standard setups used in prior distribution regression studies \cite{klein2014scalable}.

\textbf{Simulation setup.}
We run particle-mesh $N$-body simulations with grid resolution $\texttt{nc}=64$ under periodic boundary conditions.
Two simulations with different numbers of time steps are run, both evolved up to redshift $z = 0.5$:
\begin{itemize}
    \item a \emph{coarse} simulation with $4$ time steps, serving as an approximate evolution;
    \item a \emph{fine} simulation with $12$ time steps, serving as a high-fidelity reference.
\end{itemize}
Both simulations are initialized from the same initial conditions. The coarse simulation is treated as the input distribution, while the fine simulation is treated as the corresponding output distribution. \\

\textbf{Spatial binning and distribution construction.}
The spatial domain of each simulation is first rescaled to the unit cube $[0,1]^3$.
We then partition the space into a regular $16 \times 16 \times 16$ grid of cubic bins.
For each bin, we collect all particles whose spatial coordinates fall inside the bin and treat the resulting point cloud as one distribution instance.

Unless otherwise stated, we focus on the bin indexed by $(0,0,0)$, yielding paired input-output distributions corresponding to the same spatial region across simulation resolutions.
\begin{figure}[t]
    \centering
    \includegraphics[width=\linewidth]{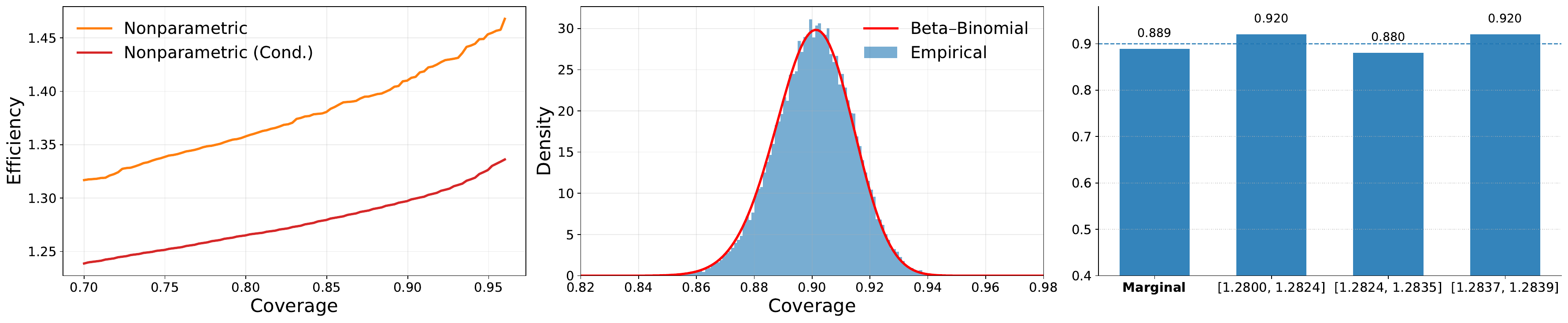}
    \caption{Experiments on Dark Matter 6D for task \textbf{(I)} (first column), \textbf{(II)} (second column), and \textbf{(III)} (third column) using the NDR base model.}
    \label{fig:DM6D}
\end{figure}
\textbf{Feature normalization.}
For each distribution instance, the spatial coordinates $(x,y,z)$ of all particles within the bin are rescaled independently to lie in $[0,1]$.
Particle velocity components $(v_x,v_y,v_z)$, when present, are left unnormalized to preserve their physical scale.

\textbf{Dimensional settings.}
We consider two variants of the Dark Matter dataset:
\begin{itemize}
    \item \textbf{3D Dark Matter:} distributions consist of spatial coordinates $(x,y,z)$ only;
    \item \textbf{6D Dark Matter:} distributions consist of spatial coordinates and velocities $(x,y,z,v_x,v_y,v_z)$.
\end{itemize}

In both cases, the input distribution is constructed from the coarse ($4$-step) simulation, and the output distribution is constructed from the fine ($12$-step) simulation.
This setup induces a challenging distribution-to-distribution regression problem, where the goal is to map approximate particle distributions to their high-fidelity counterparts.

\subsection{House Price. }\label{app: Houseprice}
\textbf{Data.  }We consider the house price dataset studied in \cite{chen2023wasserstein}. The data are based on inflation-adjusted median house prices collected from Zillow and consist of bimonthly observations for 306 cities in the United States, covering the period from June 1996 to 2025.

From the raw data, we construct the House Price dataset by extracting columns corresponding to monthly house price measurements over the period from 1996 to 2025 and retaining all rows corresponding to observed house prices across geographic locations. To ensure data integrity, any row containing a non-numeric entry is removed from the analysis. From the resulting cleaned table, we construct time-indexed input-output samples by pairing consecutive monthly columns, such that the empirical distribution of house prices at month $t$ serves as the input distribution and that at month $t+1$ serves as the corresponding output distribution. Each distribution is defined by the collection of house price observations across all retained locations. Aggregating these consecutive month pairs yields a set of distribution-to-distribution regression samples that capture temporal dynamics in housing prices. This dataset is then divided into training, testing, and calibration based on the different tasks.

\textbf{Experimental setup.}
For the House Price dataset, we adopt task-specific data splits due to the limited number of available distribution-valued observations. \\
For task \textbf{(I)}, we use $100$ distributions for training and $200$ distributions for conformal calibration. An additional $20$ distributions are reserved for cross-validation of the nonparametric regression model, and all remaining distributions are used for testing. Task-specific settings, including evaluation metrics and target coverage levels, follow the same protocol as in task \textbf{(I)} of the simulation study. For conditional Conformal Prediction, we set the neighborhood size to $k=100$. \\
For task \textbf{(III)}, we use $100$ distributions for training and $100$ distributions for calibration, with $20$ distributions reserved for cross-validation of the nonparametric regression model. The remaining distributions are used for testing. In the adaptive conformal procedure, the neighborhood size is set to $k=50$. \\
For visualization task, we use $200$ distributions for training, $100$ distributions for calibration and $20$ distributions reserved for cross-validation. \\
Task \textbf{(II)} is omitted for this dataset due to the limited amount of available data, which makes reliable evaluation infeasible. \\
All experiments are conducted using three distribution-to-distribution regression models: Nonparametric Regression, Optimal Transport Regression, and Wasserstein Regression.
\subsection{Ablation Study for Truncation Parameter}\label{app: AblationTrun}
Motivated by the importance of prediction-set volume in Conformal Prediction (see Appendix~\ref{app: backgroundConf}), we conduct an ablation study on the truncation parameter $R$, which directly controls the index set $T_R$ used in our orthonormal expansion and hence the effective dimension of the conformal set. Recall, the index set is defined as
\begin{align}
T_R \;:=\; \bigl\{ \alpha \in \mathbb{Z}^l : 0 < w_\alpha(\nu,\sigma) \le R \bigr\},
\end{align}
\begin{figure}[t]
    \centering
    \includegraphics[width=\linewidth]{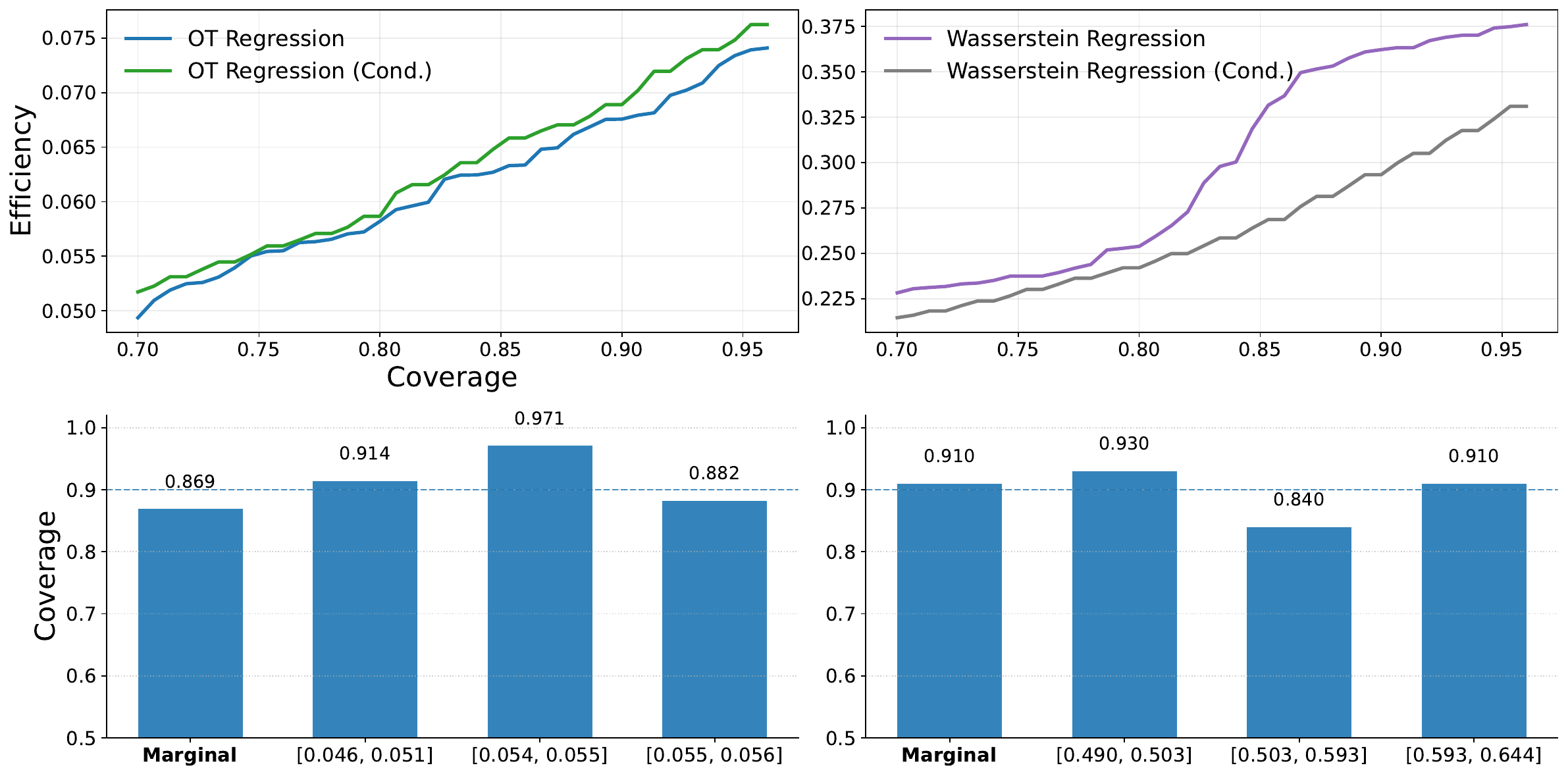}
    \caption{Experiments on House Price Data for task \textbf{(I)} (first row), and task \textbf{(III)} (second row) using the OTR base model (first column) and the WR base model (second column).}
    \label{fig:HousePriceOTRWR}
\end{figure}
and we denote its cardinality by $t := |T_R|$.

By \cref{thm: volMarg}, when $g \equiv \|\cdot\|_2$, the volume of the resulting prediction set admits the closed-form expression
\begin{align}
\mathrm{Vol}\bigl(\mathcal{Q}(X_0)\bigr)
= \frac{\pi^{t/2}}{\Gamma\!\left(\frac{t}{2}+1\right)} \cdot \frac{\hat q^{\,t}}{\sqrt{\det W}},
\label{eq:vol_exact}
\end{align}
which shows that the truncation parameter $R$ influences the prediction-set volume primarily through the induced dimension $t=|T_R|$ (and implicitly through the matrix $W$). In particular, Corollary~\ref{coro: vol} yields the asymptotic scaling
\begin{align}
\mathrm{Vol}\bigl(\mathcal{Q}(X_0)\bigr)
= O\!\left( \frac{(c\hat q)^t}{\Gamma\!\left(\frac{t}{2}+1\right)} \right),
\qquad c := \frac{\sqrt{\pi}}{m},
\label{eq:vol_asymp}
\end{align}
making explicit that the prediction-set volume depends only on the conformal quantile $\hat q$ and the truncation-induced dimension $t$.

Consequently, increasing $R$, and hence enlarging $T_R$, can rapidly inflate the prediction-set volume, reflecting the curse-of-dimensionality behavior inherent to volume-based uncertainty sets, while choosing $R$ too small may fail to capture fine-scale structure in the output distributions. 

\textbf{Experiment Setup. } 
We conduct the ablation study on the truncation parameter $R$ using the synthetic datasets considered in the simulation study, namely the Truncated Gaussian and Mixture of Betas settings. For each dataset, we evaluate three distribution-to-distribution regression models. Data splits for training, calibration, cross-validation, and testing follow the same protocol as in the simulation study, and all experimental settings are kept fixed across different values of $R$.\\
The truncation parameter is varied over the range $R \in \{2,3,...,20\}.$ For each choice of $R$, we compute the conformal quantile $\hat{q}$ from the calibration scores and use it as the primary measure of efficiency. We report $\hat{q}$ rather than the prediction-set volume, as direct volume computation becomes numerically unstable in higher dimensions, while $\hat{q}$ provides a robust and monotone proxy for prediction-set size. 

\begin{figure}[t]
    \centering
    \includegraphics[width=\linewidth]{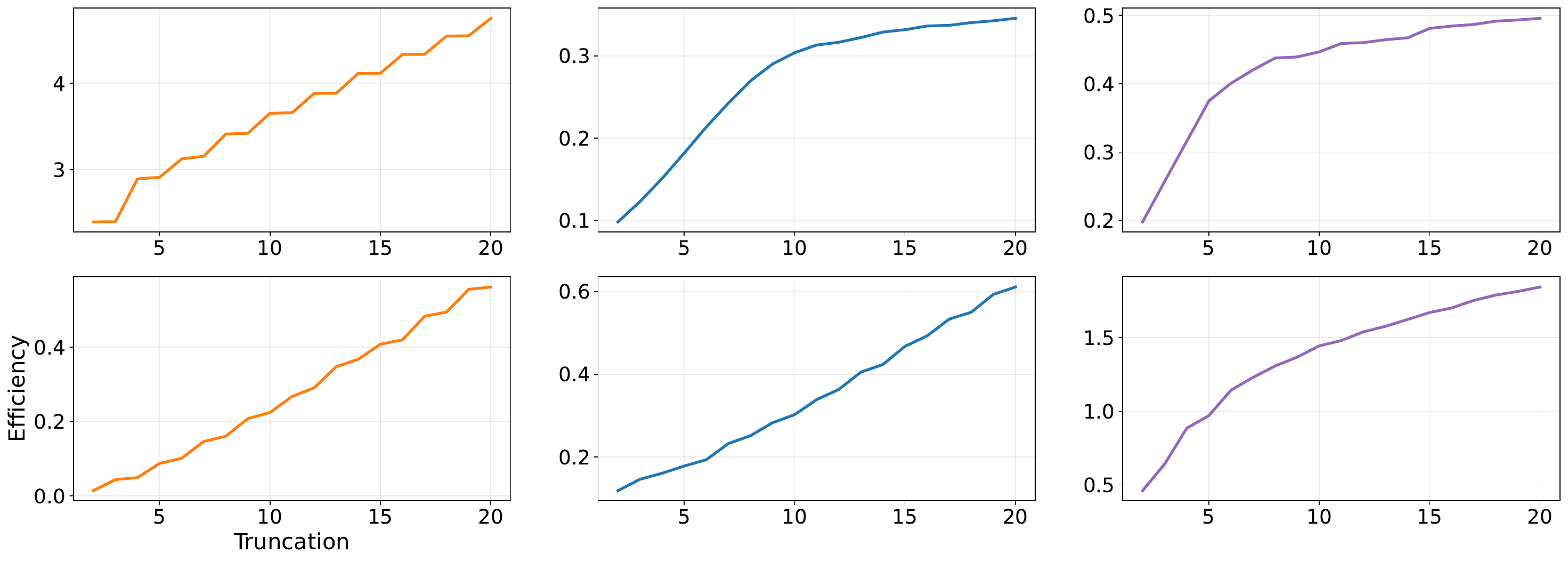}
    \caption{Ablation on the truncation parameter for Truncated Gaussian (first row) and Mixture of Betas (second row) using the NDR (first column), OTR (second column), and WR (third column) base models.}
    \label{fig:Truncation}
\end{figure}

\textbf{Results. }The results for the Truncated Gaussian and Mixture of Betas settings are reported in \cref{fig:Truncation}. Overall, the results demonstrate that computational efficiency is highly sensitive to the choice of the base regression model. Notably, even under identical data and truncation parameters in the Truncated Gaussian setting, the efficiency of Nonparametric Distribution Regression is around 3-4, whereas Optimal Transport Regression attains efficiencies on the order of 0.3-0.4, representing a more than tenfold disparity. This stark contrast empirically confirms our theoretical prediction that efficiency may grow exponentially with the truncation parameter. As a result, careful calibration of the truncation parameter to the specific model-data combination is essential in practice.
\subsection{Ablation Study for Inference Time}\label{app: AblationInftime}
While adaptive Conformal Prediction substantially reduces prediction-set size, it incurs additional computational overhead at inference time due to the localization step. In particular, for each test input distribution $X_0$, the adaptive procedure requires identifying a local calibration subset $\mathcal{C}_{X_0}$ by performing a $k$-nearest-neighbor (KNN) search in the input distribution space $(\mathcal{P}, D)$, followed by the computation of localized conformal scores and quantiles. This stands in contrast to non-adaptive Conformal Prediction, where a single global calibration set and threshold are computed once and reused for all test inputs.

More concretely, as described in \cref{sec: AdapCP}, the additional inference-time cost of adaptive Conformal Prediction arises entirely from the localization step. For a given test input distribution $X_0$, adaptive Conformal Prediction requires computing the distances.
$
\{ \mathcal D(X_0, X_i^c) \}_{i=1}^{d_0}
$
between $X_0$ and all calibration input distributions $\{X_i^c\}_{i=1}^{d_0}$ in the metric space $(\mathcal{P}, \mathcal D)$. The local index set
\begin{align}
\mathcal{I}_{X_0}
\;:=\;
\arg\min_{\substack{I \subset [n_c] \\ |I| = k}}
\sum_{i \in I} \mathcal D(X_0, X_i^c)
\end{align}
is then obtained by selecting the $k$ nearest neighbors of $X_0$ among the calibration inputs, which defines the local calibration subset $\mathcal{C}_{X_0}$. This nearest-neighbor search constitutes the dominant inference-time overhead of adaptive Conformal Prediction, with computational complexity scaling linearly with the calibration set size $d_0$ and the evaluation of the distance function $\mathcal D$.

\begin{figure}[t]
    \centering
    \includegraphics[width=\linewidth]{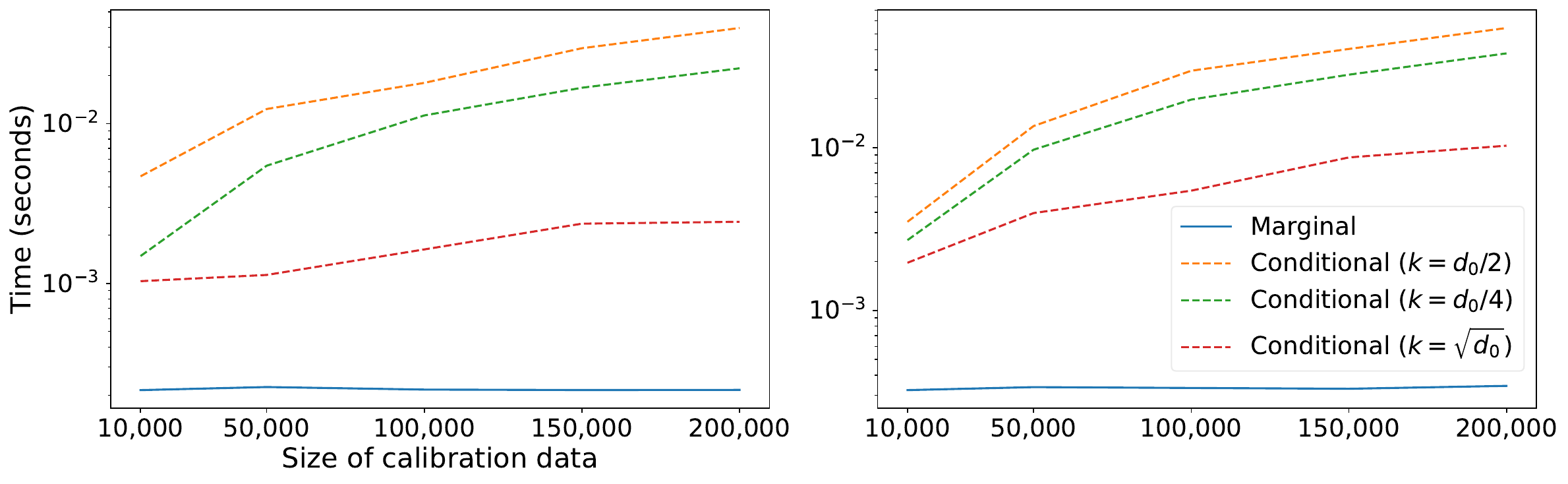}
    \caption{Ablation on Inference Time for Truncated Gaussian (first column) and Mixture of Betas (second column) using the NDR base models.}
    \label{fig:Inference}
\end{figure}

\textbf{Experiment Setup. }
We quantify the inference-time overhead of adaptive Conformal Prediction using the nonparametric regression model. This choice is motivated by the fact that inference-time complexity in conformal prediction depends only on the calibration and localization procedures, and not on the underlying regression model. \\
Experiments are conducted on the Truncated Gaussian and Mixture of Betas datasets. For each dataset, we use $1,000$ distributions for training, $200$ distributions for cross-validation, and $1,000$ distributions for testing. The size of the calibration set $d_0$ is varied over
\begin{center}
    $d_0 \in \{10,000; 50,000; 100,000; 150,000; 200,000\}$
\end{center}
To study the effect of localization on inference time, we consider multiple choices of the neighborhood size $k$. Specifically, we set
\begin{center}
    $k \in \left\{\dfrac{d_0}{2}, \dfrac{d_0}{4}, \sqrt{d_0}\right\}$
\end{center}
For each configuration, inference time is measured by applying conformal prediction to the entire test set and averaging the total runtime to obtain the mean inference time per test distribution. \\
\textbf{Results. }The results are reported in \cref{fig:Inference}. As expected, marginal Conformal Prediction exhibits nearly constant inference time, whereas adaptive Conformal Prediction incurs increasing computational overhead due to the localization step, which requires computing distances between the test input and all calibration distributions to identify the $k$ nearest neighbors. Across both datasets, inference time grows monotonically with the calibration set size and increases with $k$, confirming that the KNN search dominates the overall computational cost. While absolute runtimes differ due to dataset-specific distance metrics and data representations, the qualitative scaling behavior is consistent. Notably, adopting a sublinear neighborhood size (e.g., $k=\sqrt{d_0}$) substantially mitigates inference-time overhead, illustrating a practical trade-off between adaptivity and computational efficiency.
\subsection{Ablation Study for Hyperparameter Sensitivity}\label{app:  abonHyperparam}
Our framework introduces several hyperparameters, including the orthonormal basis $\beta$, the metric $g$ on the coefficient space, the weight matrix $W$, and the parameters $\nu,\sigma,\bar A$ from Assumption~\ref{ass: A1}. These choices are introduced both to make the theoretical analysis precise and to follow standard practice in basis-based nonparametric estimation and local conformal calibration \cite{oliva2013distribution}. A natural concern, however, is that the proposed method may be sensitive to these choices, since they affect the coefficient representation, the geometry of the prediction set, and the local calibration procedure. In this section, we empirically show that this is not the case: across a range of reasonable hyperparameter choices, our method remains stable in terms of both coverage and prediction-set size.

\textbf{Experiment setup. }To investigate hyperparameter sensitivity, we repeat the same DDR task across the [datasets] while varying one hyperparameter at a time and keeping all other settings fixed, each time we run over $5$ seeds to calculate the mean and variance. For each configuration, we measure the empirical coverage and compare it with the nominal target level. For the orthonormal basis $\beta$, we replace the default basis with alternative standard choices, including the Legendre basis and a wavelet basis. For the regularity parameters, we vary $\nu \in \{1,1.5,2\}$ and $\sigma \in \{0.5,1,1.5,2\}$. For the score function, we consider $g = \|\cdot\|_1$ and $g = \|\cdot\|_3$. For the truncation level, we test $R \in \{3,5,8\}$, and for the neighborhood size, we test $k \in \{250,500,750\}$. We do not vary $W$ or $\bar A$ in this experiment because the weight matrix $W$ is fully determined by the choices of $\nu$ and $\sigma$, while $\bar A$ is a fixed regularity constant used in the theoretical assumption and does not enter the implementation of the conformal prediction procedure.

\textbf{Results. }The results are reported in Table~\ref{tab:hyperparam_sensitivity}. Overall, across both the Truncated Gaussians and Mixture of Betas datasets, the empirical coverage remains consistently close to the nominal level, with only minor variation across hyperparameter choices. This indicates that the proposed method is not overly sensitive to hyperparameter selection and remains robust under reasonable perturbations of its main design parameters.
\begin{table}[t]
\centering
\caption{Ablation on Hyperparameter Sensitivity results.}
\label{tab:hyperparam_sensitivity}
\begin{tabular}{lcc}
\toprule
\textbf{Varied Parameter} & \textbf{Truncated Gaussians} & \textbf{Mixture of Betas} \\
\midrule
Legendre Basis & $0.90 \pm 0.013$ & $0.90 \pm 0.014$ \\
Wavelet Basis  & $0.90 \pm 0.013$ & $0.90 \pm 0.013$ \\
\midrule
$\nu = 1$   & $0.90 \pm 0.014$ & $0.90 \pm 0.014$ \\
$\nu = 1.5$ & $0.90 \pm 0.013$ & $0.90 \pm 0.013$ \\
$\nu = 2$   & $0.90 \pm 0.013$ & $0.90 \pm 0.013$ \\
\midrule
$\sigma = 0.5$ & $0.90 \pm 0.013$ & $0.90 \pm 0.013$ \\
$\sigma = 1$   & $0.90 \pm 0.014$ & $0.90 \pm 0.014$ \\
$\sigma = 1.5$ & $0.90 \pm 0.013$ & $0.90 \pm 0.013$ \\
$\sigma = 2$   & $0.90 \pm 0.014$ & $0.90 \pm 0.013$ \\
\midrule
$g = \|\cdot\|_3$ & $0.90 \pm 0.014$ & $0.90 \pm 0.013$ \\
$g = \|\cdot\|_1$ & $0.90 \pm 0.013$ & $0.90 \pm 0.013$ \\
\midrule
$R = 3$ & $0.90 \pm 0.013$ & $0.90 \pm 0.014$ \\
$R = 5$ & $0.90 \pm 0.013$ & $0.90 \pm 0.013$ \\
$R = 8$ & $0.90 \pm 0.014$ & $0.90 \pm 0.013$ \\
\midrule
$k = 250$ & $0.87 \pm 0.007$ & $0.89 \pm 0.016$ \\
$k = 500$ & $0.88 \pm 0.009$ & $0.89 \pm 0.018$ \\
$k = 750$ & $0.89 \pm 0.010$ & $0.90 \pm 0.019$ \\
\bottomrule
\end{tabular}
\end{table}
\subsection{Low-Data Regime Analysis}\label{app: lowdata}
Our method is built upon Split Conformal Prediction, which constructs prediction sets using a dedicated calibration split. While this design enables finite-sample coverage guarantees, it also increases data requirements compared to standard training-testing pipelines. This limitation becomes particularly pronounced in low-data settings, where a small number of samples must simultaneously support model training, calibration, and evaluation. 

To investigate the behavior of our approach under such constraints, we evaluate performance in a low-data regime using the Mortality dataset of \cite{ghodrati2022distribution}, which contains only 36 samples. This dataset represents an extreme yet practically relevant scenario in which both the base regression model and the conformal calibration procedure operate under severe data scarcity. In this low-data setting, we focus our evaluation on task \textbf{(I)}, and task \textbf{(III)} (recall the definitions in \cref{sec: Exp} and Appendix~\ref{App: Símtudy}). These tasks directly probe the reliability and adaptivity of the conformal prediction sets under data scarcity. We do not consider task \textbf{(II)}, since accurate estimation of the underlying Beta distribution requires a sufficiently large sample size, which is not available in this regime.

\textbf{Experimental setup. } Experiments use the same three distribution-to-distribution regression models as in previous sections. \\
For task \textbf{(I)}, the data are split so that 10 distributions are used to fit the base regression model, while 20 distributions are reserved for conformal calibration. Cross-validation for the nonparametric regression model is performed using 2 additional distributions, and all remaining distributions are reserved for evaluation. The conditional conformal procedure is applied with a neighborhood size of $k=10$. Since the available data are extremely limited, we repeat this random partitioning $20$ times and report averages over the resulting runs.\\
For task \textbf{(III)}, the same data split is used, with distributions assigned to training, calibration, tuning, and testing as described above. The adaptive conformal procedure is implemented with a neighborhood size $k=10$. All results are reported as averages over 5 independent runs. \\
For the visualization task, the data are split into 20 distributions for training, 2 distributions for cross-validation, and 10 distributions for calibration.

\begin{figure}[t]
    \centering
    \includegraphics[width=\linewidth]{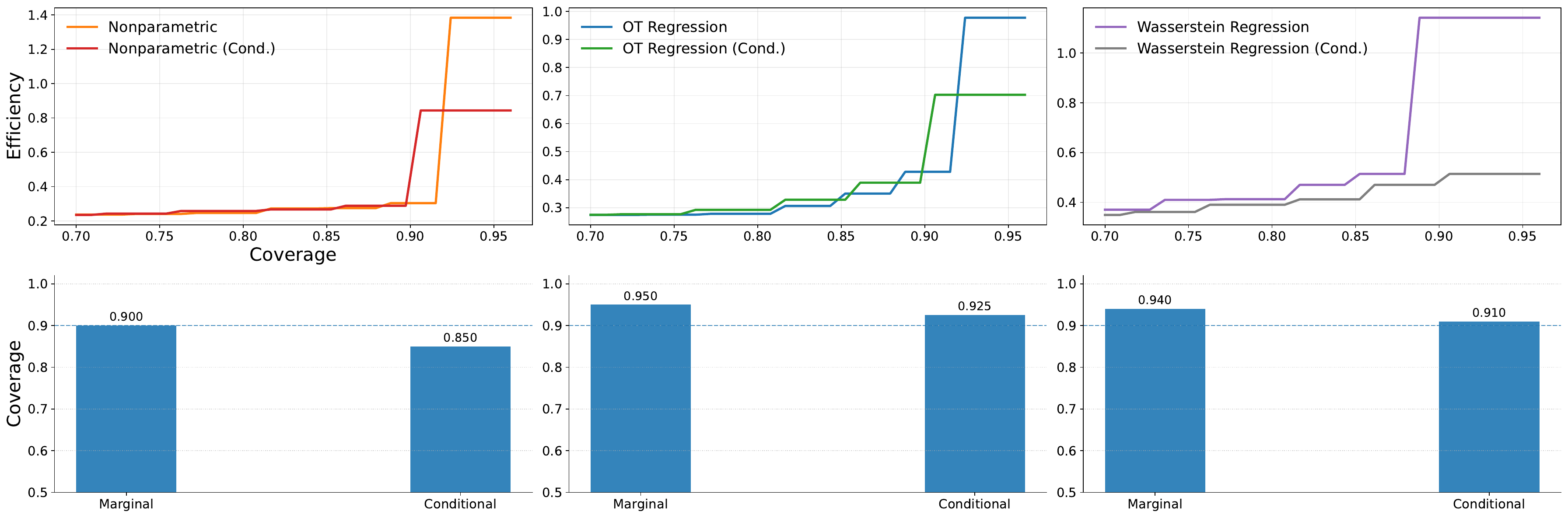}
    \caption{Ablation in the low-data regime on the Mortality dataset for task \textbf{(I)} (first row) and task \textbf{(III)} (third row) using the NDR (first column), OTR (second column), and WR (third column) base models.}
    \label{fig:Morality}
\end{figure}

\textbf{Results. }The results are shown in \cref{fig:Morality}. For task \textbf{(I)}, the limited sample size leads to a noisier and less smooth efficiency-coverage curve; nevertheless, the overall trend remains consistent with that observed in larger-scale experiments, and the adaptive variant continues to produce prediction sets that are comparable in size or smaller than those of the non-adaptive method. For task \textbf{(III)}, adaptive conformal prediction achieves coverage levels close to both the marginal baseline and the nominal target, demonstrating that the proposed method remains effective even in the low-data regime.

\textbf{Visualization. }To further illustrate the effectiveness of our method in the low-data regime, we uniformly sample distributions from the prediction set and overlay them with the true density of the dataset. The resulting visualizations are shown in \cref{fig: Dist Est}.

\subsection{Computational Infrastructure}
All experiments are conducted on a CPU-only environment. In the simulation study, generating the Truncated Gaussian and Mixture of Betas datasets takes about 1 minute, while generating the Dark Matter data requires running a numerical simulation and takes on the order of one day. \\
For the experimental tasks, the end-to-end runtimes (including model training, calibration, and evaluation) are as follows. Task \textbf{(I)} takes roughly 2 hours, task \textbf{(II)} takes around 3 hours, and task \textbf{(III)} takes about 1 hour. The visualization task requires around 10 minutes. The ablation study on the truncation parameter takes about 30 minutes, while the ablation study for inference time takes roughly 5 hours.

\subsection{Broader Societal Impacts} \label{app: Broader Societal Imapcts}
This work is primarily methodological and theoretical. Its potential positive societal impact lies in improving uncertainty quantification for distribution-valued prediction tasks, where both inputs and outputs may represent complex empirical distributions rather than finite-dimensional observations. Such settings arise in a range of scientific, medical, environmental, and policy-relevant applications, where reliable uncertainty estimates are important for interpreting predictions and supporting downstream decisions. By developing calibrated prediction sets for distribution-to-distribution regression, the proposed framework may help practitioners assess not only a single predicted distribution but also a statistically valid range of plausible output distributions. This can make uncertainty more transparent in settings where distributional outcomes are central, such as population-level health analysis, demographic forecasting, climate or environmental modeling, economic risk assessment, and scientific simulation. The ability to characterize prediction-set volume and sample from the resulting sets may further support visualization, sensitivity analysis, and decision-making under uncertainty.

%% file: checklist.tex
\section*{NeurIPS Paper Checklist}

\begin{enumerate}

\item {\bf Claims}
    \item[] Question: Do the main claims made in the abstract and introduction accurately reflect the paper's contributions and scope?
    \item[] Answer: \answerYes{} 
    \item[] Justification: The claims made in the abstract and introduction accurately reflect the paper's theoretical contributions and scope. They are supported by the main theorems and are stated under explicit assumptions. 

\item {\bf Limitations}
    \item[] Question: Does the paper discuss the limitations of the work performed by the authors?
    \item[] Answer: \answerYes{} 
    \item[] Justification: Limitations of this work is discussed in Section~\ref{sec: Conc}. 

\item {\bf Theory assumptions and proofs}
    \item[] Question: For each theoretical result, does the paper provide the full set of assumptions and a complete (and correct) proof?
    \item[] Answer: \answerYes{} 
    \item[] Justification: We do provide the full set of assumptions and complete proofs for the main theorems. 

    \item {\bf Experimental result reproducibility}
    \item[] Question: Does the paper fully disclose all the information needed to reproduce the main experimental results of the paper to the extent that it affects the main claims and/or conclusions of the paper (regardless of whether the code and data are provided or not)?
    \item[] Answer: \answerYes{} 
    \item[] Justification: We fully disclose all the information needed to reproduce the main experimental results of the paper to the extent that it affects the main claims and conclusions of the paper. 

\item {\bf Open access to data and code}
    \item[] Question: Does the paper provide open access to the data and code, with sufficient instructions to faithfully reproduce the main experimental results, as described in supplemental material?
    \item[] Answer: \answerYes{} 
    \item[] Justification: We provide open access to the data and code in Supplementary Material, with sufficient instructions to faithfully reproduce the main experimental results. 

\item {\bf Experimental setting/details}
    \item[] Question: Does the paper specify all the training and test details (e.g., data splits, hyperparameters, how they were chosen, type of optimizer) necessary to understand the results?
    \item[] Answer: \answerYes{} 
    \item[] Justification: We specify all the training and test details (e.g., data splits, hyperparameters, how they were chosen, type of optimizer) necessary to understand the results in Appendix~\ref{app: expdet}. 

\item {\bf Experiment statistical significance}
    \item[] Question: Does the paper report error bars suitably and correctly defined or other appropriate information about the statistical significance of the experiments?
    \item[] Answer: \answerYes{} 
    \item[] Justification: The paper reports suitably and correctly defined error bars. 

\item {\bf Experiments compute resources}
    \item[] Question: For each experiment, does the paper provide sufficient information on the computer resources (type of compute workers, memory, time of execution) needed to reproduce the experiments?
    \item[] Answer: \answerYes{} 
    \item[] Justification: We provide sufficient information on the computer resources (type of compute workers, memory, time of execution) needed to reproduce the experiments. 
    
\item {\bf Code of ethics}
    \item[] Question: Does the research conducted in the paper conform, in every respect, with the NeurIPS Code of Ethics \url{https://neurips.cc/public/EthicsGuidelines}?
    \item[] Answer: \answerYes{} 
    \item[] Justification: The research conducted in the paper conforms with the NeurIPS Code of Ethics. 

\item {\bf Broader impacts}
    \item[] Question: Does the paper discuss both potential positive societal impacts and negative societal impacts of the work performed?
    \item[] Answer: \answerYes{} 
    \item[] Justification: Broader Societal Impacts are discussed in Appendix~\ref{app: Broader Societal Imapcts}.

\item {\bf Safeguards}
    \item[] Question: Does the paper describe safeguards that have been put in place for responsible release of data or models that have a high risk for misuse (e.g., pre-trained language models, image generators, or scraped datasets)?
    \item[] Answer: \answerNA{} 
    \item[] Justification: The paper does not release any data, pre-trained models, image generators, or scraped datasets. Therefore, safeguards for responsible release are not applicable.  

\item {\bf Licenses for existing assets}
    \item[] Question: Are the creators or original owners of assets (e.g., code, data, models), used in the paper, properly credited and are the license and terms of use explicitly mentioned and properly respected?
    \item[] Answer: \answerYes{} 
    \item[] Justification:  The paper has credited all code and data that have been used. 

\item {\bf New assets}
    \item[] Question: Are new assets introduced in the paper well documented and is the documentation provided alongside the assets?
    \item[] Answer: \answerNA{} 
    \item[] Justification: The paper does not introduce new assets such as datasets, trained models, or software packages. Therefore, asset documentation is not applicable. 

\item {\bf Crowdsourcing and research with human subjects}
    \item[] Question: For crowdsourcing experiments and research with human subjects, does the paper include the full text of instructions given to participants and screenshots, if applicable, as well as details about compensation (if any)? 
    \item[] Answer: \answerNA{} 
    \item[] Justification: The paper does not involve crowdsourcing experiments or research with human subjects. Therefore, participant instructions, screenshots, and compensation details are not applicable. 

\item {\bf Institutional review board (IRB) approvals or equivalent for research with human subjects}
    \item[] Question: Does the paper describe potential risks incurred by study participants, whether such risks were disclosed to the subjects, and whether Institutional Review Board (IRB) approvals (or an equivalent approval/review based on the requirements of your country or institution) were obtained?
    \item[] Answer: \answerNA{} 
    \item[] Justification: The paper does not involve study participants or human-subject research. Therefore, participant risks, risk disclosure, and IRB approval are not applicable. 

\item {\bf Declaration of LLM usage}
    \item[] Question: Does the paper describe the usage of LLMs if it is an important, original, or non-standard component of the core methods in this research? Note that if the LLM is used only for writing, editing, or formatting purposes and does \emph{not} impact the core methodology, scientific rigor, or originality of the research, a declaration is not required.
    \item[] Answer: \answerNA{} 
    \item[] Justification: LLM is used
only for writing, editing, or formatting purposes and does not impact the core methodology,
scientific rigor, or originality of the research.

\end{enumerate}